\documentclass{article}

\PassOptionsToPackage{numbers, compress}{natbib}
\usepackage[preprint]{neurips_2023}

\usepackage[utf8]{inputenc} % allow utf-8 input
\usepackage[T1]{fontenc}    % use 8-bit T1 fonts
\usepackage{hyperref}       % hyperlinks
\usepackage{url}            % simple URL typesetting
\usepackage{booktabs}       % professional-quality tables
\usepackage{amsfonts}       % blackboard math symbols
\usepackage{nicefrac}       % compact symbols for 1/2, etc.
\usepackage{microtype}      % microtypography
\usepackage{xcolor}         % colors

\usepackage{color}
\usepackage{graphicx}
\usepackage{wrapfig}
\usepackage{picinpar}

\usepackage{times}
\usepackage{soul}
\usepackage{url}
\usepackage[small]{caption}
\usepackage{graphicx}
\usepackage{amsmath}
\usepackage{amsthm}
\usepackage{booktabs}
\usepackage{algorithm}
\usepackage{algorithmic}

\usepackage{amssymb, bm}		% Mine
\usepackage{wrapfig}		% Mine
\usepackage{subfigure}		% Mine
\usepackage{color}			% Mine
\usepackage{xcolor}			% Mine
\usepackage{multicol}		% Mine
\usepackage{multirow} 		% Mine
\usepackage{bbding}
\usepackage[figuresright]{rotating}
\usepackage{verbatim}
\usepackage{siunitx}

\usepackage{adjustbox}		% Mine
\usepackage{booktabs}		% Mine
\usepackage{makecell}
\usepackage{bbm}

\DeclareMathOperator*{\argmax}{arg\,max}

\newcommand{\bs}{\mathbf{s}}
\newcommand{\ba}{\mathbf{a}}
\newcommand{\btau}{\mathbf{\tau}}

\newcommand{\bx}{\mathbf{x}}
\newcommand{\by}{\mathbf{y}}
\newcommand{\bz}{\mathbf{z}}

\newcommand{\eg}{\emph{e.g.},\ }
\newcommand{\ie}{\emph{i.e.},\ }

\newtheorem{theorem}{Theorem}%[section]
\newtheorem{lemma}[theorem]{Lemma}

\title{Beyond OOD State Actions: Supported Cross-Domain Offline Reinforcement Learning}

\author{%
Jinxin Liu\thanks{Equal contributions. \ \  $^{1}$Westlake University. \ \  Correspondence to: <wangdonglin@westlake.edu.cn>.}$^{\ \ 1}$
\And 
Ziqi Zhang$^{* 1}$
\And 
Zhenyu Wei$^{1}$
\And 
Zifeng Zhuang$^{1}$
\AND 
Yachen Kang$^{1}$
\And 
Sibo Gai$^{1}$
\And 
Donglin Wang$^{1}$
}

\begin{document}

\maketitle

\begin{abstract}
Offline reinforcement learning (RL) aims to learn a policy using only pre-collected and fixed data. Although avoiding the time-consuming online interactions in RL, it poses challenges for out-of-distribution (OOD) state actions and often suffers from data inefficiency for training. Despite many efforts being devoted to addressing OOD state actions, the latter (data inefficiency) receives little attention in offline RL. To address this, this paper proposes the cross-domain offline RL, which assumes offline data incorporate additional source-domain data from varying transition dynamics (environments), and expects it to contribute to the offline data efficiency. To do so, we identify a new challenge of OOD transition dynamics, beyond the common OOD state actions issue, when utilizing cross-domain offline data. Then, we propose our method BOSA, which employs two support-constrained objectives to address the above OOD issues. Through extensive experiments in the cross-domain offline RL setting, we demonstrate BOSA can greatly improve offline data efficiency: using only 10\% of the target data, BOSA could achieve {74.4\%} of the SOTA offline RL performance that uses 100\% of the target data. Additionally, we also show BOSA can be effortlessly plugged into model-based offline RL and noising data augmentation techniques (used for generating source-domain data), which naturally avoids the potential dynamics mismatch between target-domain data and newly generated source-domain data. 
\end{abstract}

\section{Introduction}

\begin{wrapfigure}{r}{0.525\textwidth}
	\vspace{-12pt}
	\begin{center}
		\includegraphics[scale=0.475]{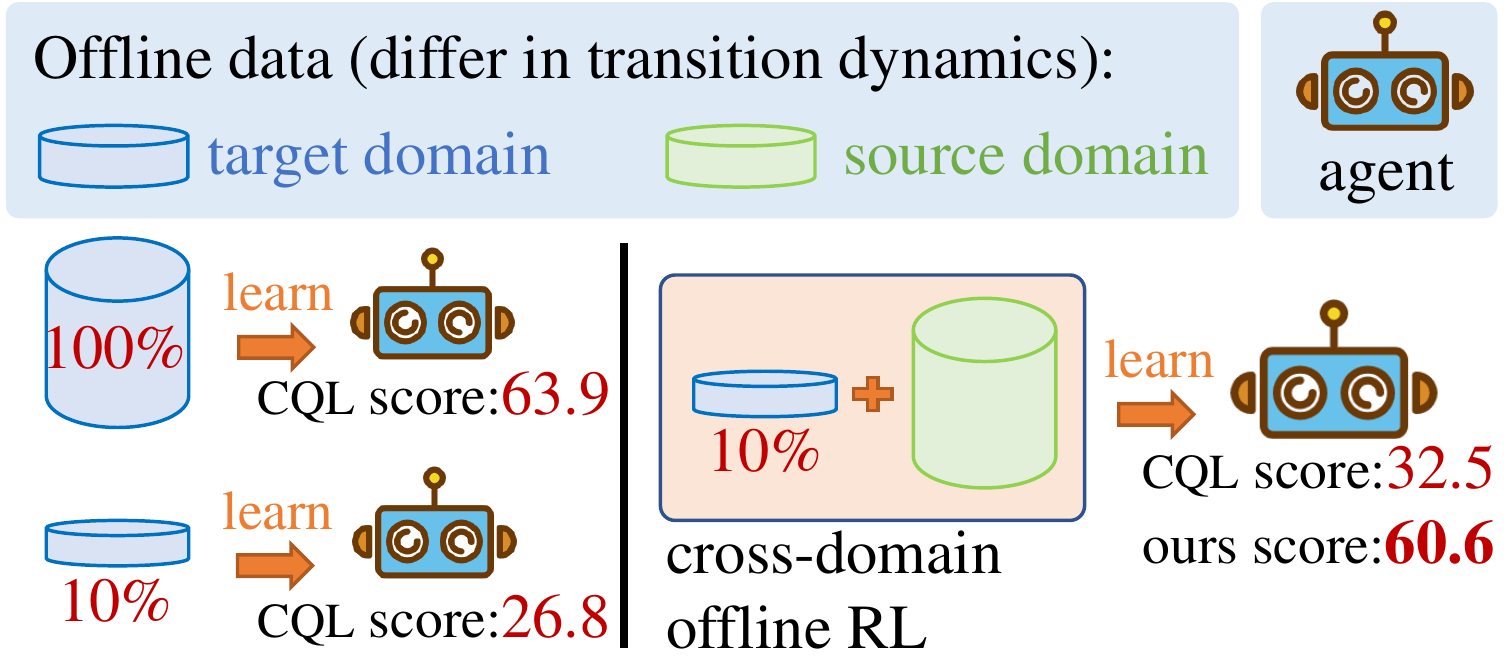}
	\end{center}
	\vspace{-8pt}
	\caption{(\textit{left}) The vanilla sing-domain offline RL setting. (\textit{right})~Our cross-domain offline RL, which learns a policy from offline cross-domain data, consists of \textit{limited target-domain data} and {plenty of source-domain data}. Scores in this diagram are averaged over 9 D4RL Mujoco tasks, which serve as the target domain.
	}
	\vspace{-16pt}
	\label{fig:bosa-framework}
\end{wrapfigure}

\begin{figure*}[t]
	%	\vspace{-2pt}
	\begin{center}
		\includegraphics[scale=0.20]{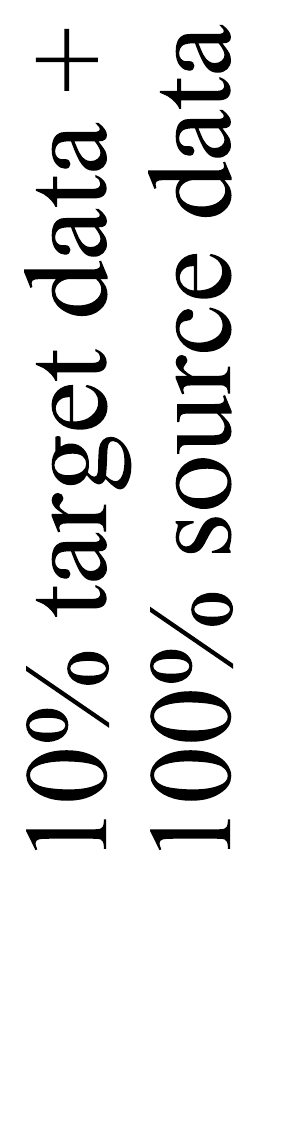}
		\includegraphics[scale=0.20]{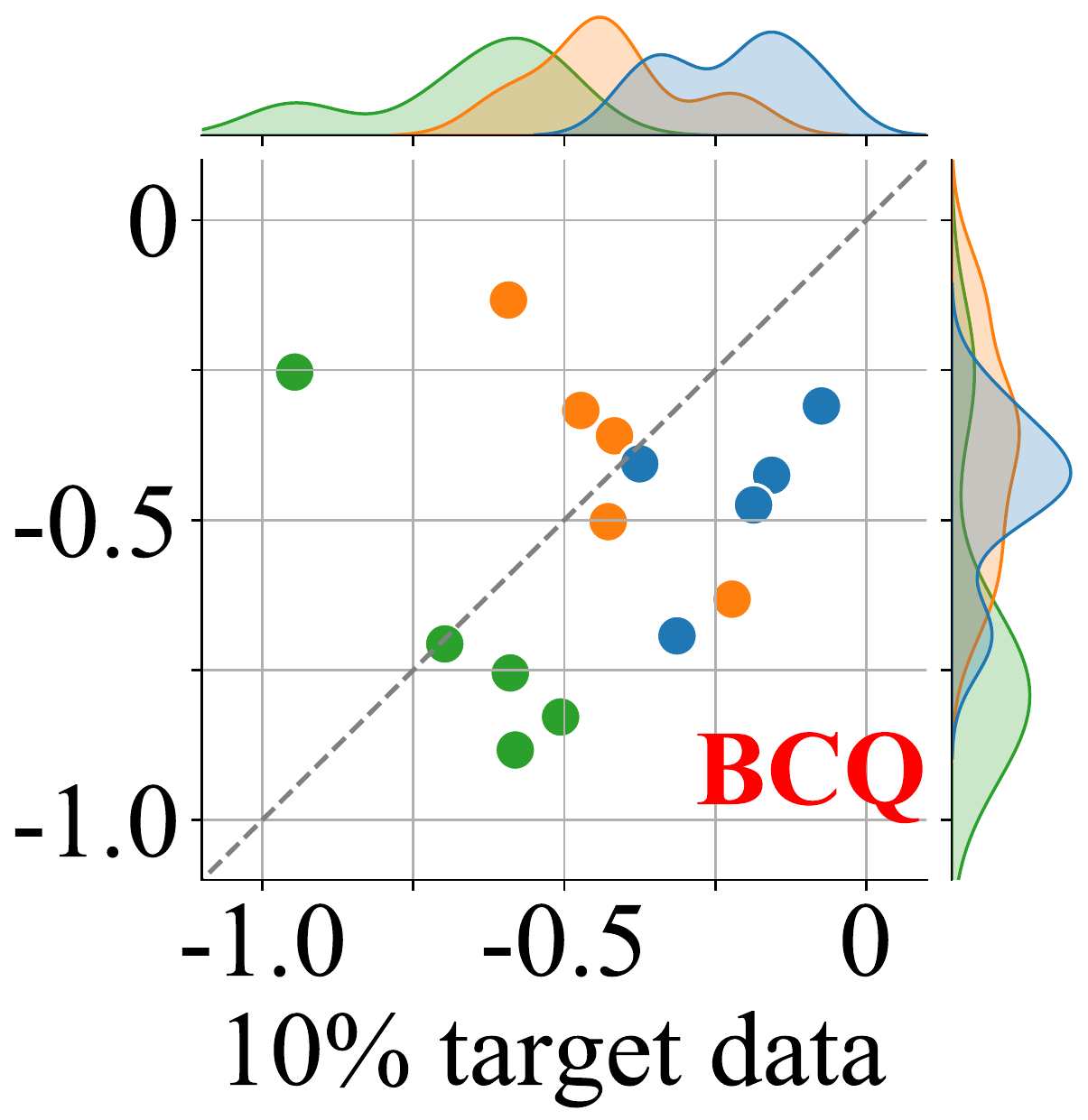}
		\includegraphics[scale=0.20]{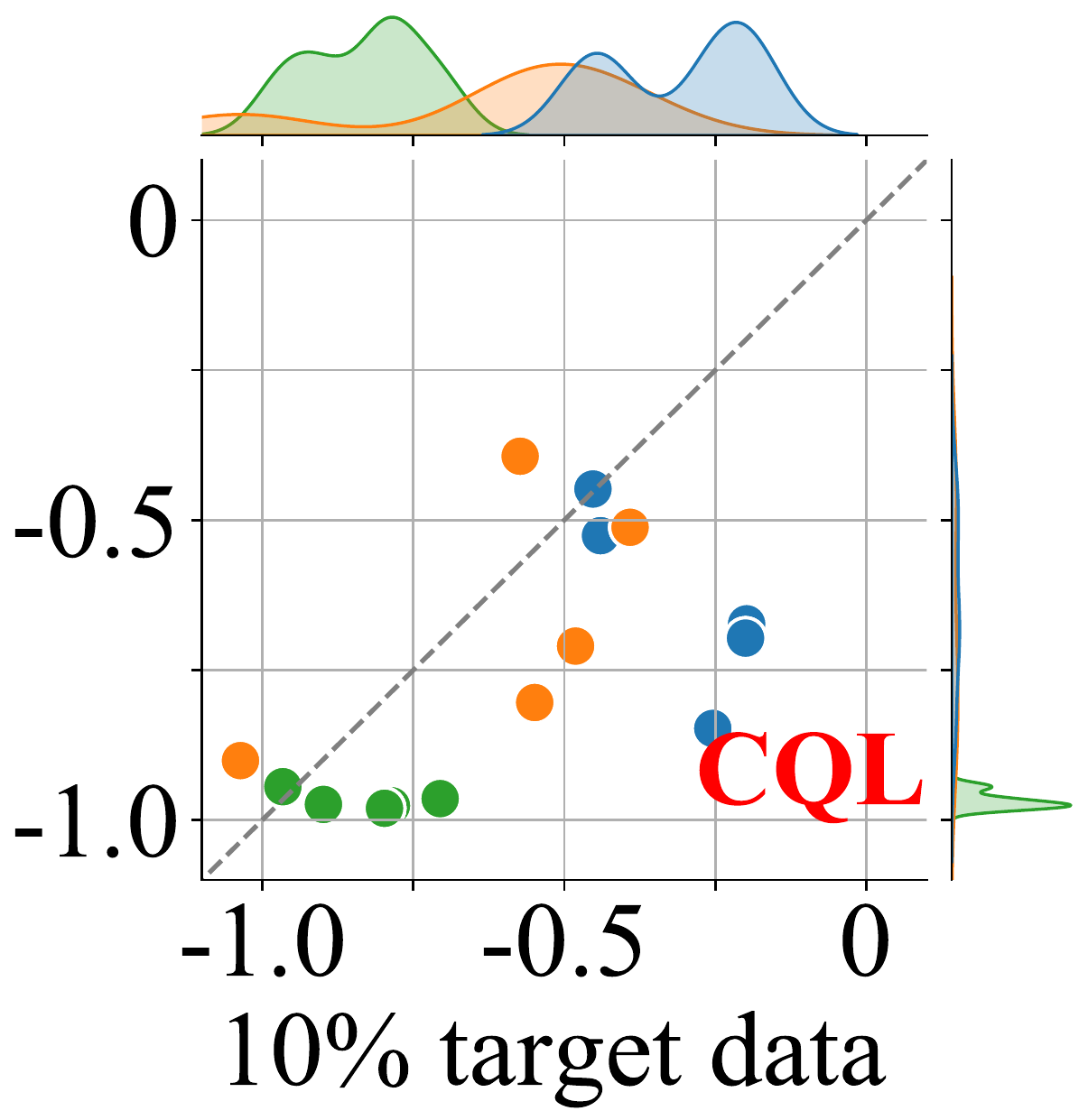}
		\includegraphics[scale=0.20]{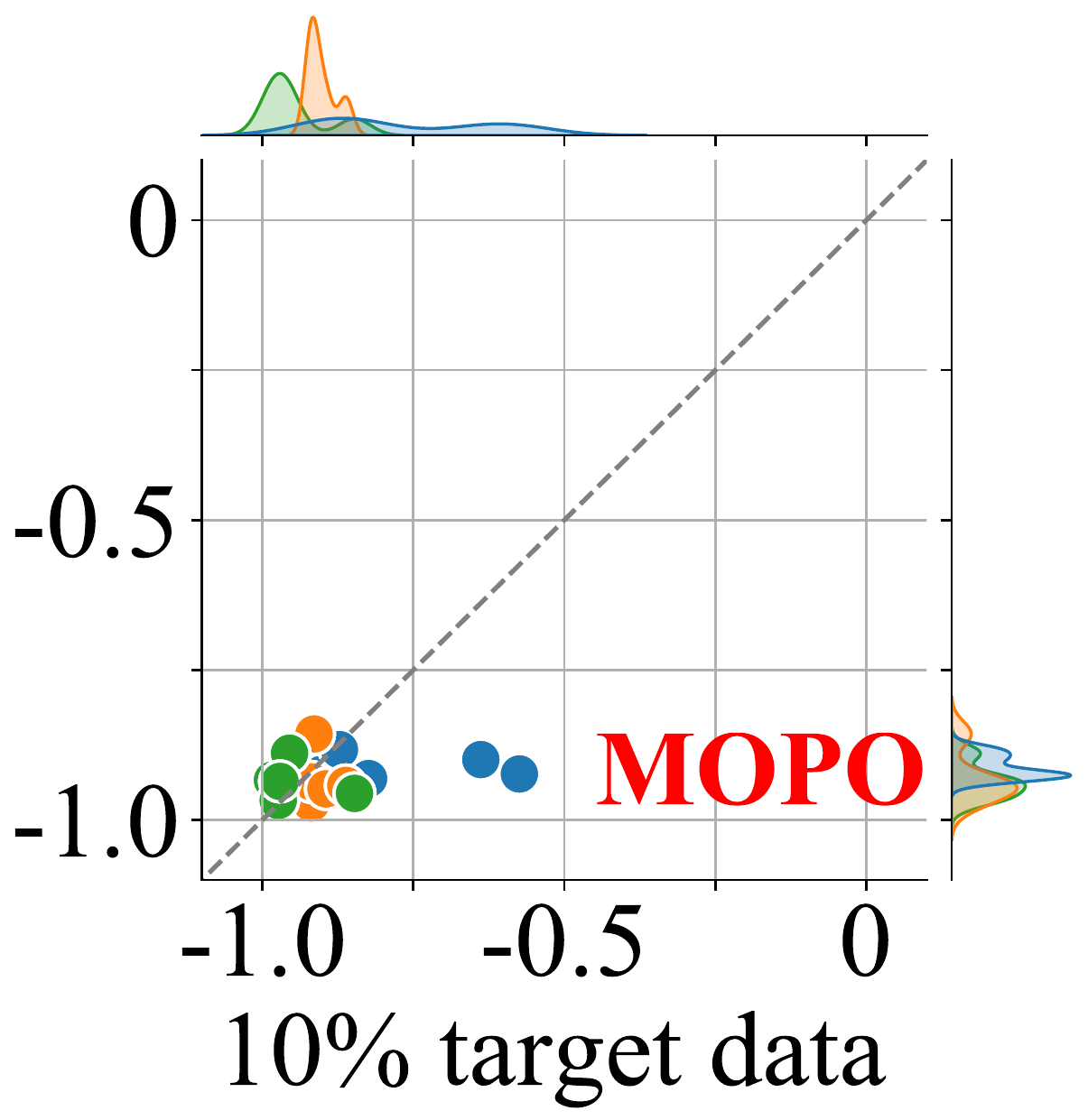}
		\includegraphics[scale=0.20]{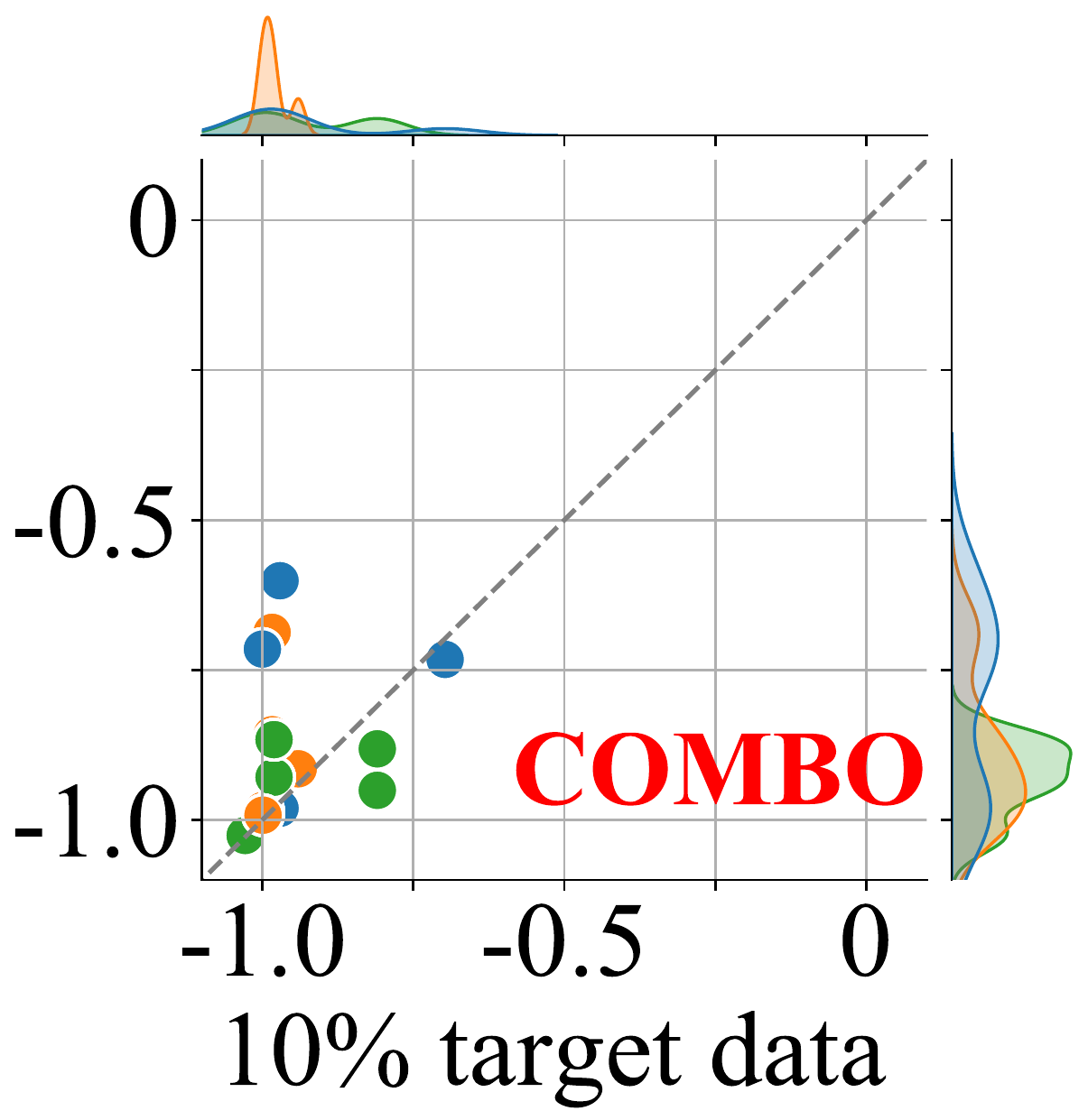}
		\includegraphics[scale=0.20]{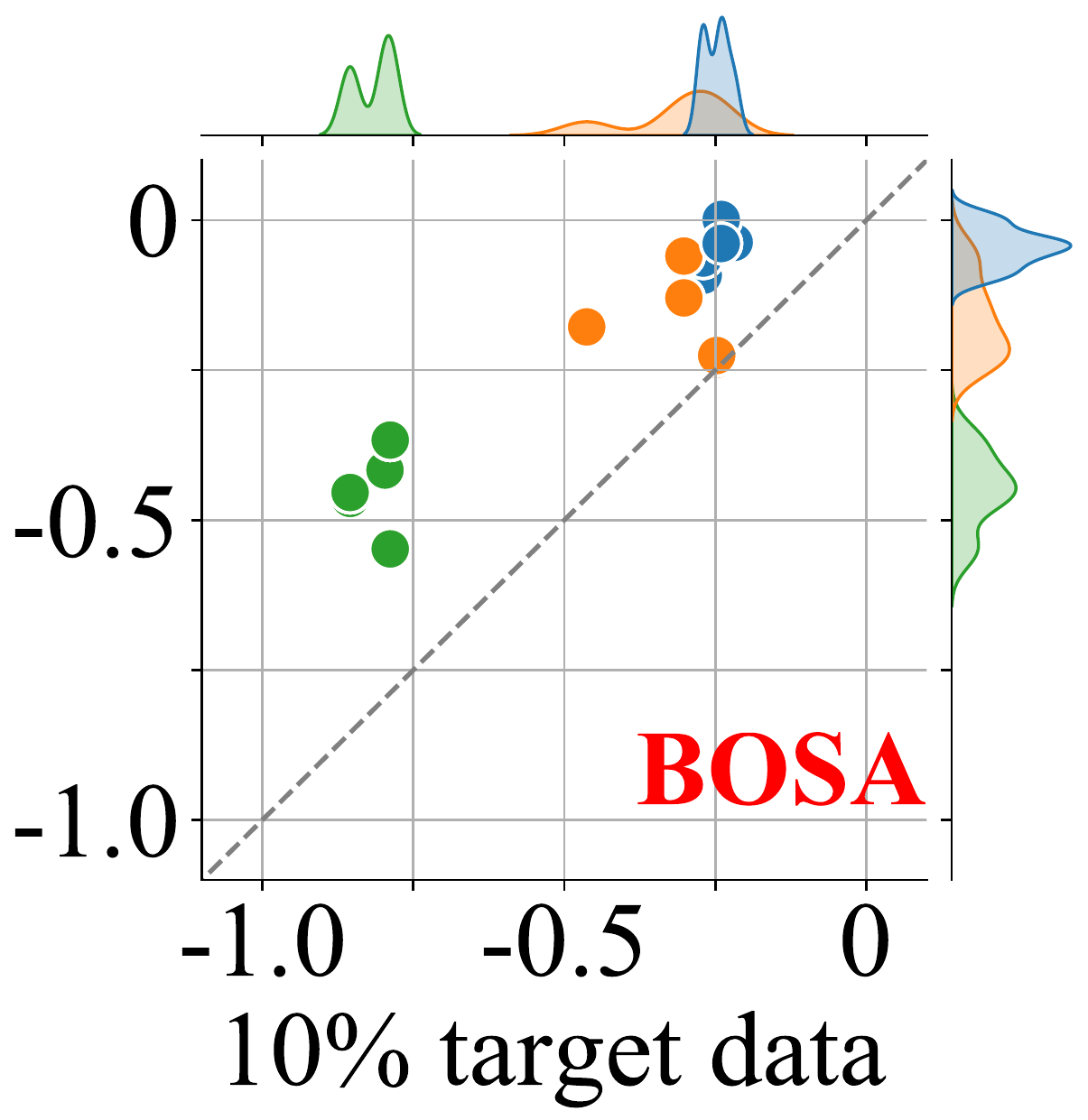}
	\end{center}
	\vspace{-9pt}
	\caption{Performance difference between single-domain and cross-domain offline RL settings, where different colors represent different data qualities (blue: *-medium, green: *-medium-replay, red: *-medium-expert) and multiple dots with the same color represent scores on multiple cross-domain tasks. 
    We take D4RL~\cite{d4rl2020Fu} as the target domain, and take a modified D4RL (with transition dynamics shift) as the source domain. The x-axis represents the normalized performance improvement when only 10\% of D4RL data (target) is available, and the y-axis represents the performance difference between learning with cross-domain offline data (10\%target + 100\%source) and learning with abundant target-domain offline data (100\%target). Formally, $\text{x} = \frac{\text{Score}(\text{10\%target})-\text{BestScore}(\text{100\%target})}{\text{BestScore}(\text{100\%target})}$, and  $\text{y} = \frac{\text{Score}(\text{10\%target + 100\%source})-\text{BestScore}(\text{100\%target})}{\text{BestScore}(\text{100\%target})}$. We can observe that when we reduce the offline training data size, most offline RL methods suffer a clear drop in performance (\ie values on the x-axis is less than 0). Further, introducing additional source-domain data also does not bring any significant performance benefits (\ie below the dashed line), with the exception of our BOSA. 
    % (m: medium, mr: medium-replay, me: medium-expert)
    }
	\label{fig:intro-vs}
	\vspace{-12pt}
\end{figure*}

Data-driven offline reinforcement learning (RL) holds the promise to learn a control policy from fixed and static dataset~\cite{levine2020offline} while avoiding the costly and time-consuming online data acquisition in standard RL. 
Nevertheless, a notorious challenge in offline RL is the presence of extrapolation error, which tends to drive the learning policy towards OOD state actions and yields collapsed behaviors~\cite{fujimoto2019off}. 
Many recent works have been dedicated to tackling this challenge, leveraging policy regularization, value conservation, or supervised regression~\cite{chen2021decision,fujimoto2021minimalist,kostrikov2021offlineiql}.
Yet, such offline solutions tend to perform well when plenty of pre-collected training data is available and the testing environment keeps consistent with the data-collecting environment. However, collecting large-scale offline data for a domain-specific environment is still labor-intensive and expensive. %, especially for real-world offline tasks. 
Additionally, once we reduce the offline training data, the final performance drops quickly (Figure~\ref{fig:bosa-framework} \textit{left} and Figure~\ref{fig:intro-vs}). 
Therefore, although that advanced offline RL methods tend to be saturated with abundant data, the underlying offline data inefficiency makes them difficult to use for data-scarce scenarios, especially for real-world offline~tasks.

Motivated by the sample-efficient domain adaptation, this paper investigates the cross-domain offline RL.  Specifically, we assume the agent can access a large amount of source-domain offline data from one or multiple separate source environments, and we are interested in adapting the learning policy with limited target-domain offline data (as shown in Figure~\ref{fig:bosa-framework} \textit{right}). We also emphasize that such an additional source-domain data assumption can be naturally satisfied in practice. For example, it is common to have a large amount of data on driving behaviors in city roads (source domain), while only having a few samples in mountain environments (target domain) for autonomous~driving~tasks.

However, prior offline RL methods often work poorly in the cross-domain setting, particularly for source-domain data with large transition dynamics differences. As we show in Figure~\ref{fig:intro-vs}, simply combing the cross-domain data does not bring a positive  performance improvement to that using only the limited target data. In fact, it can even result in a decline in performance for several tasks and offline baselines. In this paper, we attribute this transfer failure to the presence of OOD transition dynamics beyond those (OOD state actions) commonly encountered in single-domain offline RL setting. Intuitively, the presence of source-domain data can bias the agent's policy to visit transitions in source environments as long as they receive high values. 
Thus, cross-domain data easily lead the policy overfits to the source environment and hinders its transfer ability to the target domain. 

To address this issue, we present BOSA (\textbf{B}eyond \textbf{O}OD \textbf{S}tate \textbf{A}ctions) for the cross-domain offline RL. Simply, BOSA aims to leverage cross-domain offline data (plenty of source data and limited target data) to improve the data efficiency for offline RL. The key idea behind BOSA is that we substantiate the inherent offline extrapolation error through OOD state actions and OOD transition dynamics, and try to filter out offline transitions that might cause state-actions shift and transition dynamics mismatch. Specifically, we propose a supported policy optimization for eliminating OOD state actions and a supported value optimization for addressing OOD transition dynamics. Additionally, to avoid exploiting overestimated Q-values for policy optimization over source-domain data, we introduce a conservation~\cite{kumar2020conservative} over the value optimization objective. 

We conduct experiments with a variety of source domains that have dynamics mismatch and demonstrate that BOSA contributes to significant gains on learning from cross-domain data. Further, we show that BOSA can be plugged into more general cross-domain offline settings: model-based RL and (noising) data augmentation. Similarly, augmenting the offline data by a pseudo-transition model or random noise will also encounter OOD transitions that are not consistent with the target environment. Thus, we can naturally view the generated or noised data as a source domain, and then apply BOSA for handling the potential dynamics mismatch. 

The primary contributions of this work are as follows: 
1) We identify an OOD transition dynamics issue in cross-domain offline RL and propose BOSA for handling it. 
2) We show BOSA can greatly improve offline data efficiency and outperform prior state-of-the-art methods. 
3) We show BOSA can be naturally plugged into model-based RL and (noising) data augmentation scenarios while eliminating the commonly overlooked~OOD transition dynamics and thus facilitating positive transfer.

\section{Related Work}
\textbf{Offline Reinforcement Learning.} 
In standard (single domain) offline RL, the agent tries to learn a policy from static and fixed data that are pre-collected from a (target) environment~\cite{fujimoto2019off,levine2020offline}. Yet, the overestimation of OOD state-action issues is often identified as a major issue. To solve this, most model-free offline methods typically augment existing off-policy algorithms with a penalty~\cite{geist2019theory} measuring a divergence between the learning policy and the offline data (or the behavior policy). 
Recently, there have been many offline RL methods proposed to implement such a penalty, by introducing support constrains~\cite{fujimoto2019off,ghasemipour2021emaq,wu2022supported} or policy regularization~\cite{fujimoto2021minimalist,kang2023beyond,kostrikov2021offlineiql,kumar2019stabilizing,lai2023chipformer,nachum2019algaedice,peng2019advantage,wu2019behavior,zhuang2023behavior}.  
Alternatively, some offline RL methods introduce the uncertainty estimation \cite{an2021uncertainty,bai2022pessimistic,ma2021conservative,rezaeifar2021offline,wu2021uncertainty} or the conservation \cite{kumar2020conservative,liu2023drop,lyu2022mildly} over values to overcome the potential overestimation for OOD state actions. 
In the same spirit, model-based RL methods similarly employ the distribution % (correcting)
regularization~\cite{hishinuma2021weighted,yang2022regularizing,zhang2022discriminator}, uncertain estimation~\cite{yu2020mopo,kidambi2020morel,lu2021revisiting}, and value conservation~\cite{yu2021combo} to eliminate OOD issues.

\textbf{Cross-Domain Reinforcement Learning.}
To improve the sample efficiency, cross-domain RL introduces an additional source domain and regards the original task as the target domain. 
In this work, we assume source and target domains differ in their transition dynamics.  
This cross-dynamics setting has also helped improve the training efficiency for online RL~\cite{eysenbach2020off,eysenbach2021mismatched,wulfmeier2017mutual}, reward-free RL~\cite{chebotar2021actionable,liu2021unsupervised,liu2022learn,tian2021unsupervised}, and reverse RL (imitation learning)~\cite{fickinger2021cross,franzmeyer2022learn,kang2021off,viano2021robust}. 
In the offline RL setting, our BOSA method bears a resemblance to that in DARA~\cite{liu2022dara}, but a crucial algorithmic difference is that \textit{DARA explicitly models both the source-domain and target-domain transitions} while \textit{we only model target-domain transitions and do not model the source-domain transitions}, and also lead to improved performance empirically.

\section{Background}
\textbf{Reinforcement Learning.}  
We consider reinforcement learning (RL) in the Markov decision process (MDP) defined by a tuple $\mathcal{M}:=(\mathcal{S}, \mathcal{A}, T, r, \gamma, p_0)$, where $\mathcal{S}$ denotes the state space, $\mathcal{A}$ denotes the action space, $T: \mathcal{S} \times \mathcal{A} \times \mathcal{S} \to \mathbb{R}_{+}$ is the transition (dynamics) probability, $r: \mathcal{S} \times \mathcal{A} \to \mathbb{R}$ is the reward function, $\gamma \in [0, 1]$ is the discount factor, and $p_0$ is the distribution of initial states. 
The goal of RL is to find a policy $\pi: \mathcal{S} \times \mathcal{A} \to \mathbb{R}_{+}$ that maximizes the expected discounted cumulative return $J(\pi) = \mathbb{E}_{\btau\sim\pi(\btau)}\left[ \sum_{t=0}^{T} \gamma^t r_t \right]$, where $\btau:=(\bs_0, \ba_0, \bs_1, \ba_1 \cdots)$ denotes the rollout trajectory and $r_t:=r(\bs_t, \ba_t)$ denotes the reward of transition $(\bs_t, \ba_t )$ at time step $t$. Here we sightly abuse the notation $\pi(\btau)$ to denote the trajectory distribution induced by executing the policy $\pi(\ba|\bs)$ in $\mathcal{M}$, \ie $\bs_0 \sim p_0(\bs_0)$, $\ba_t \sim \pi(\ba_t|\bs_t)$, and $\bs_{t+1} \sim T(\bs_{t+1}|\bs_t, \ba_t)$. 

To optimize the above objective $J(\pi)$, off-policy RL methods introduce a Q-function $Q^\pi(\bs, \ba)$ defined by $Q^\pi(\bs, \ba) = \mathbb{E}_{\btau\sim\pi(\btau)}\left[ \sum_{t=0}^{T} \gamma^t r_t \vert \bs_0=\bs_t, \ba_0=\ba \right]$. One property of such a Q-value function is that it satisfies the Bellman consistency criterion given by $\mathcal{T}^{\pi}Q^\pi(\bs, \ba) = Q^\pi(\bs,\ba)$ $\forall (\bs, \ba)$, where $\mathcal{T}^{\pi} Q(\bs, \ba):= r(\bs, \ba) + \gamma \mathbb{E}_{\bs'\sim T(\bs'|\bs,\ba), \ba'\sim\pi(\ba'|\bs')}$ $\left[ Q(\bs',\ba') \right]$ is the Bellman operator. 
Given an experience replay buffer $\mathcal{B}:=\{(\bs,\ba,r,\bs')\}$ (that will be updated by executing the learning policy in the environment), standard approximate dynamic programming and actor-critic methods use this (Bellman consistency) criterion to iteratively learn a parametric Q-function $Q_\phi$ by minimizing, % the mean squared Bellman error, 
\begin{align}
	\mathcal{L}_\mathcal{B}(Q_\phi) := \mathbb{E}_{(\bs, \ba, \bs')\sim\mathcal{B}} \left[ Q_\phi(\bs, \ba) - \mathcal{T}^{\pi_\theta} Q_{\bar{\phi}}(\bs,\ba) \right]^2, \label{eq:policy-evaluation}
\end{align}
%\begin{align}
%\mathbb{E}_{(\bs, \ba, r, \bs')\sim\mathcal{B},\ba'\sim\pi_\theta(\ba'|\bs')} \left[ Q_\phi(\bs, \ba) - r - Q_{\bar{\phi}}(\bs', \ba')  \right]^2, \label{eq:policy-evaluation}
%\end{align}
and, following the deterministic policy gradient theorem, learn a parametric policy $\pi_\theta$ by maximizing:
\begin{align}
	\mathcal{J}_\mathcal{B}(\pi_\theta) := \mathbb{E}_{\bs \sim \mathcal{B}, \ba\sim\pi_\theta(\ba|\bs)}\left[ Q_\phi(\bs, \ba) \right], \label{eq:policy-improvement}
\end{align}
where $\phi$ and $\theta$ are the parameters of the Q-function and the policy respectively, 
$\bar{\phi}$ is an EMA (exponential moving average) of $\phi$: $\bar{\phi} \leftarrow \alpha \bar{\phi} + (1-\alpha)\phi$, and $\alpha$ is the target network EMA parameter. For simplicity of notation, we drop the subscript $t$ and use $\bs'$ to denote the state at the next time step. 

%Note that in standard online RL setting, 

\textbf{Offline Reinforcement Learning.} 
In offline RL~\cite{levine2020offline}, we can not execute the learning policy in the environment to collect new online transitions, but rather have access to a fixed offline dataset $\mathcal{D}:=\{(\bs,\ba,r,\bs')\}$, collected by an unknown behavior policy (or by multiple behaviors) $\pi_\beta(\ba|\bs)$ in the environment $\mathcal{M}$. %    \sim \text{Rollout}(\pi_\beta, \mathcal{M})
%Then, we introduce 
%the empirical Bellman operator $\hat{\mathcal{B}}_{\pi}Q$, 
%\begin{align}\label{eq:empirical-bellman-operator}
%\hat{\mathcal{B}}_{\pi} Q(\bs, \ba)= r + \gamma \mathbb{E}_{\bs'\sim \hat{T}(\bs'|\bs,\ba), \ba'\sim\pi(\ba'|\bs')} \left[ Q(\bs',\ba') \right],
%\end{align}
%where $\hat{T}:=\argmax_{\hat{T}}\mathbb{E}_{(\bs,\ba,\bs')\sim\mathcal{D}}\left[\log \hat{T}(\bs'|\bs,\ba)\right]$ is an empirical estimation on the offline transition dynamics in $\mathcal{D}$. 
%However, naively querying the empirical Bellman operator $\hat{\mathcal{B}}_{\pi_\theta}Q_{\bar{\phi}}$ (as a target for training $Q_\phi$) will 

However, naively performing policy evaluation (Equation~\ref{eq:policy-evaluation}) %  and policy improvement (Equation~\ref{eq:policy-improvement}) 
and taking the expectation over fixed offline data $\mathcal{D}$ will inevitably require the Q-function to extrapolate to OOD state-action pairs. Iterating the offline policy improvement and policy evaluation \ie $\max \mathcal{J}_\mathcal{D}(\pi_\theta)$ and $\min \mathcal{L}_\mathcal{D}(Q_\phi)$, the potential extrapolation error will be further amplified, biasing the learned Q-value towards erroneously overestimated values and further biasing the learned policy towards unconfident actions. 
Unlike the online RL formulation, such an induced extrapolation error and the unconfident action will never be corrected due to the inability to collect new interaction data over the task environment.

\section{Problem Formulation}

\subsection{Cross-Domain Offline RL} 
\textbf{Problem statement.} 
In our cross-domain offline RL setting, we assume the static offline data are collected from a set of environments/MDPs with varying transition dynamics (and different data-collecting behavior policies), rather than from a single fixed environment like the vanilla single-domain offline RL formulation in~\citet{levine2020offline}. 

%{we assume that offline data are \textit{not} collected from a single environment}. Rather, we assume the cross-domain offline data are collected from a set of environments with varying transition dynamics and different behavior policies. 

Formally, considering a target offline RL task specified through $\mathcal{M}_\text{target}$, we define the mixed cross-domain offline data $\mathcal{D}_\text{mix} := \mathcal{D}_\text{target} \cup \mathcal{D}_\text{source}$, where $\mathcal{D}_\text{target}$ denotes the (limited) target data collected from the target MDP $\mathcal{M}_\text{target}$ and $\mathcal{D}_\text{source}$ denotes the source data collected from a set of source MDPs $\{\mathcal{M}^1_\text{source}, \cdots, \mathcal{M}^n_\text{source}\}$. We also assume that all of these MDPs share the same state space, action space, and reward function, while differing in their transition dynamics. The goal of cross-domain offline RL is to learn a policy that maximizes the expected return over {the target environment} $\mathcal{M}_\text{target}$ using  the static cross-domain offline data $\mathcal{D}_\text{mix}$. 

Compared to the vanilla single-domain offline RL, the cross-domain formulation naturally preserves the benefit of offline data transfer. Incorporating the source-domain offline data $\mathcal{D}_\text{source}$ can alleviate the challenges of offline RL data efficiency, which often requires a large number of target offline samples and demands substantial data collection efforts on the target environment. % especially in real-world offline tasks. 
Thus, we expect that cross-domain offline RL formulation can reduce the demanding requirements on target data. 

%To instantiate this cross-domain setting, we introduce three different strategies for introducing the new source domain data over the vanilla single-domain offline RL: (1) Following the dynamics-transfer setting, we can assume 

\subsection{OOD Issues in Cross-Domain Offline RL}
Before discussing the OOD issues in the cross-domain setting, we first review the extrapolation error and take a deeper look at how to eliminate it with support constraints.

%review the OOD {state actions issue} in the vanilla (single-domain) offline RL and take a deeper look at how to eliminate it with support constraints. 

% \subsubsection{OOD State Actions} 
\textbf{OOD state actions.}  
In typical single-domain offline RL (learning with offline data $\mathcal{D}$), performing the offline policy evaluation will suffer from the empirical extrapolation error $\mathbb{E}_{\mu_\pi(\bs)\pi(\ba|\bs)}\vert \epsilon(\bs,\ba) \vert$, where $\epsilon(\bs,\ba) = \mathcal{T}^{\pi}Q(\bs,\ba) - \hat{\mathcal{T}}^{\pi}Q(\bs,\ba)$, and $\hat{\mathcal{T}}$ denotes the empirical Bellman operator~\cite{fujimoto2019off} implicitly defined by the offline transition dynamics $\hat{T}$ by randomly sampling transitions $(\ba,\bs,r,\bs')$ from $\mathcal{D}$.  

Thus, to evaluate a policy $\pi$ exactly, we are required to ensure $\mathbb{E}_{\mu_\pi(\bs)\pi(\ba|\bs)}\vert \epsilon(\bs,\ba) \vert=0$ {at relevant state-action transitions}. For this purpose, BCQ~\cite{fujimoto2019off} identifies OOD state actions as the key source of the extrapolation error and thus proposes the following theorem. 
\begin{theorem}
	\label{theorm:ood-state actions}
	Under deterministic environment dynamics, if we can ensure all possible state actions are contained in offline data $\mathcal{D}$, we can guarantee $T(\bs'|\bs,\ba) - \hat{T}(\bs'|\bs,\ba)=0$ for all $\bs' \in \mathcal{S}$ and $(\bs,\ba)$ such that $\mu_\pi(\bs)\pi(\ba|\bs) > 0$. Then, $\mathbb{E}_{\mu_\pi(\bs)\pi(\ba|\bs)}\vert \epsilon(\bs,\ba) \vert=0$ will be naturally satisfied. 
\end{theorem}
Based on such a theorem, BCQ suggests that one can eliminate the extrapolation error by instantiating offline RL over a support-constrained paradigm \textit{which constraints the learned policy (actions) within the support set of the offline dataset}~\cite{fujimoto2021minimalist, ghasemipour2021emaq}. 

% \footnote{This analysis further inspires many successful methods for support-constrained policy evaluation/improvement~\cite{ghasemipour2021emaq,fujimoto2021minimalist}}, 

\textbf{OOD transition dynamics.} 
We can observe that the above support-constrained derivation relies on the identification that %OOD state actions servers as the key source of extrapolation error. 
if we can ensure $(\bs, \pi_\theta(\bs)) \in \mathcal{D}$ for all $\bs \in \mathcal{D}$, then we can achieve zero extrapolation error. %  $T(\bs'|\bs,\ba) - \hat{T}(\bs'|\bs,\ba)=0$ for all $\bs' \in \mathcal{S}$ (Theorem~\ref{theorm:bcq}). 
However, such an identification is only limited to the single-domain offline RL. The main reason is that in the cross-domain setting, it is easy to find a transition $(\bs,\ba,\bs')$ such that $(\bs,\ba,\bs') \in \mathcal{D}_\text{mix}$ and $T_\text{target}(\bs'|\bs,\ba) \neq \hat{T}_\text{mix}(\bs'|\bs,\ba)$. Thus, we can not guarantee $T_\text{target}(\bs'|\bs,\ba) - \hat{T}_\text{mix}(\bs'|\bs,\ba)=0$ for all $(\bs,\ba)\in\mathcal{D}_\text{mix}$ and $\bs' \in \mathcal{S}$ like Theorem~\ref{theorm:ood-state actions}. 
Even though we can restrict state actions $(\bs, \pi_\theta(\bs))$ to lie in the support set of the offline data $\mathcal{D}_\text{mix}$, performing policy evaluation will still accumulate non-zero extrapolation errors due to the transition dynamics mismatch (between the target and source MDPs).  % $\epsilon_\text{mix}(\bs,\ba) := \mathcal{T}_\text{target}^{\pi}Q(\bs,\ba) - \hat{\mathcal{T}}_\text{mix}^{\pi}Q(\bs,\ba) \neq 0$. 

\begin{lemma}
	\label{lemma:ood-transitions}
	Under the cross-domain offline RL setting,  
	%performing support-constrained policy improvement over $\mathcal{D}_\text{mix}$ (\ie 
	constraining the policy within the support of cross-domain data $\mathcal{D}_\text{mix}$ can not guarantee zero extrapolation error for the target environment when performing offline policy evaluation. 
\end{lemma}

%We can also define the target-relevant extrapolation error $\mathbb{E}_{\mu_\pi(\bs)\pi(\ba|\bs)}\vert \epsilon_\text{mix}(\bs,\ba) \vert$, where $\epsilon_\text{mix}(\bs,\ba) := \mathcal{T}_\text{target}^{\pi}Q(\bs,\ba) - \hat{\mathcal{T}}_\text{mix}^{\pi}Q(\bs,\ba) $ 

Thus, beyond the common OOD state actions issue identified in previous offline works, cross-domain offline RL also suffers from {OOD transition dynamics} (transition dynamics mismatch). In the next section, we will describe how our method, BOSA, addresses both of these issues jointly by introducing supported policy and value optimization. %incorporating support constraints into

\section{Supported Policy and Value Optimization}

In this section, we present BOSA (beyond OOD state actions). As aforementioned, cross-domain offline RL suffers from both OOD state actions and OOD dynamics issues. BOSA tackles these two issues jointly: considering the actor-critic framework, BOSA eliminates OOD state actions through a supported policy optimization (Section~\ref{sec:supported-po}) and addresses OOD dynamics through a supported value optimization (Section~\ref{sec:supported-vo}). In Section~\ref{sec:implementation}, we then present a practical implementation of BOSA.

\subsection{Supported Policy Optimization}
\label{sec:supported-po}

Following the same spirit of BCQ~\cite{fujimoto2019off}, we first introduce the supported policy optimization to address the OOD state actions issue. 
Note that the naive BCQ algorithm formulates the supported policy optimization over the Q-learning algorithm, utilizing the \textit{optimal} Bellman operator. Here we rewrite it on top of the actor-critic methods, separating the policy improvement and the policy evaluation (value optimization). Specifically, considering a parametric Q-function $Q_\phi$ and a policy network $\pi_\theta$, we can %similarly perform the offline policy evaluation by minimizing $\mathcal{L}_\mathcal{B}(Q_\phi)$ (replacing the expectation distribution in Equation~\ref{eq:policy-evaluation} with offline data $\mathcal{D}$) while 
perform offline \textit{support-constrained} policy optimization by 
\begin{align}
	\max_{\pi_\theta} \ \mathcal{J}_{\mathcal{D}_\text{mix}}(\pi_\theta) &:= \mathbb{E}_{\bs \sim \mathcal{D}_\text{mix}, \ba\sim\pi_\theta(\ba|\bs)}\left[ Q_\phi(\bs, \ba) \right], 
	\text{s.t.} \ (\bs, \pi_\theta(\bs)) \in \mathcal{D}_\text{mix}, \  \forall \bs \in \mathcal{D}_\text{mix}, \label{eq:bcq-ac-policy-support}
\end{align} 
where Equation~\ref{eq:bcq-ac-policy-support} constraints the learned policy (actions) within the support set of the offline dataset, thus eliminating the OOD state actions issue as specified by Theorem~\ref{theorm:ood-state actions}. 

Unfortunately, directly optimizing Equation~\ref{eq:bcq-ac-policy-support} is often computationally expensive and requires a tabular representation for the state actions, which can quickly become impractical for large problems. Instead, we approximate it through an alternative objective by using a behavior policy $\hat{\pi}_{\beta_\text{mix}}$: % w.r.t the offline data $\mathcal{D}_\text{mix}$: 
\begin{align}
	\max_{\pi_\theta} \ \mathcal{J}_{\mathcal{D}_\text{mix}}(\pi_\theta) &:= \mathbb{E}_{\bs \sim \mathcal{D}_\text{mix}, \ba\sim\pi_\theta(\ba|\bs)}\left[ Q_\phi(\bs, \ba) \right],  
	\text{s.t.} \ \mathbb{E}_{\bs\sim\mathcal{D}_\text{mix}}\left[ \log \hat{\pi}_{\beta_\text{mix}}(\pi_\theta(\bs)|\bs) \right] > \epsilon_\text{th}, \label{eq:bosa-policy-support}
\end{align} 
where $\hat{\pi}_{\beta_\text{mix}} = \argmax_{\beta_\text{mix}}\mathbb{E}_{(\bs,\ba)\sim\mathcal{D}_\text{mix}}\left[ \log \hat{\pi}_{\beta_\text{mix}}(\ba|\bs) \right]$ and $\epsilon_\text{th}$ denotes the threshold above which we retain the (state-action) support constrains. 
Note that compared to the common policy distribution matching regularization in prior offline methods (\eg restricting $\text{KL}(\pi_\theta(\ba|\bs)||\pi_\beta(\ba|\bs)) \leq \epsilon_\text{th}$), Equation~\ref{eq:bosa-policy-support} is essentially performing support constraints instead of distribution matching, which thus avoids the difficulty to trade off mode-covering and mode-seeking issues in distribution matching~objective.

\subsection{Supported Value Optimization}
\label{sec:supported-vo}

Next, we discuss how to tackle OOD transition dynamics. Similar to the above support-constrained policy optimization, the key idea is to constrain the value optimization (policy evaluation) over the transitions that do not expose the dynamics mismatch. 
As an example, one can directly use only the target-domain offline data $\mathcal{D}_\text{target}$ to perform value optimization by minimizing $\mathcal{L}_\text{target}(Q_{{\phi}})$, where $\mathcal{L}_\text{target}(Q_{{\phi}}):= \mathbb{E}_{(\bs, \ba, r, \bs')\sim\mathcal{D}_\text{target},\ba'\sim\pi_\theta(\ba'|\bs')} \left[ Q_\phi(\bs, \ba) - r - Q_{\bar{\phi}}(\bs', \ba')  \right]^2$. 
However, this naive method tends to suffer from low data efficiency and struggles with scarce (target-domain) offline data, especially in data-expensive offline RL tasks. 

To facilitate data-efficient value optimization, leveraging the source-domain data $\mathcal{D}_\text{source}$ is thus essential in the cross-domain setting. Following the same spirit of policy support constraints in Equation~\ref{eq:bosa-policy-support}, we introduce the supported value optimization: 
\begin{align}
	\min_{Q_\phi} \ \mathcal{L}_\text{mix}(Q_{{\phi}}):= \mathbb{E}_{^{(\bs, \ba, r, \bs')\sim\mathcal{D}_\text{mix}}_{\ba'\sim\pi_\theta(\ba'|\bs')}} \Big[ \Big. 
	\left. \delta(Q_\phi) \cdot \mathbbm{1}( \hat{T}_\text{target}(\bs'|\bs,\ba) > \epsilon_\text{th}')\right] + \mathbb{E}_{(\bs,\ba) \sim \mathcal{D}_\text{source}}\left[ Q_\phi(\bs,\ba) \right]\label{eq:bosa-value-support}
\end{align}

%\begin{align}
%	\min_{Q_\phi} \ \mathcal{L}_\text{mix}(Q_{{\phi}}):= &\mathbb{E}_{(\bs, \ba, r, \bs')\sim\mathcal{D}_\text{mix},\ba'\sim\pi_\theta(\ba'|\bs')} \Big[ \Big. \nonumber\\
%&\left. \delta(Q_\phi) \cdot \mathbbm{1}(\log \hat{T}_\text{target}(\bs'|\bs,\ba) > \epsilon_\text{th}')\right] \label{eq:bosa-value-support}\\
%	&+ \mathbb{E}_{(\bs,\ba) \sim \mathcal{D}_\text{source}}\left[ Q_\phi(\bs,\ba) \right] 
%	\label{eq:bosa-value-support}
%\end{align}
where $\delta(Q_\phi):=\left( Q_\phi(\bs, \ba) - r - Q_{\bar{\phi}}(\bs', \ba')  \right)^2$, $\hat{T}_\text{target}$ denotes the estimated target-domain transition dynamics, \ie $\hat{T}_\text{target} = \argmax_{\hat{T}_\text{target}}\mathbb{E}_{(\bs,\ba,\bs')\sim\mathcal{D}_\text{target}}\left[ \log {\hat{T}_\text{target}}(\bs'|\bs, \ba) \right]$, and $\mathbbm{1}(\cdot)$ denotes the indicator function, which equals $1$ if the argument is true, and $0$ otherwise. 
Intuitively, the indicator function filters out OOD transitions that are likely to yield dynamics mismatching. 
Further, we also introduce a conservative regularization in Equation~\ref{eq:bosa-value-support} that encourages learning conservative Q values for source-domain data, thus avoiding exploiting false and overestimated values when performing policy optimization in Equation~\ref{eq:bosa-policy-support}.

\textbf{Comparison to dynamics-aware reward modification.} 
We also note that recent works propose to learn a dynamics-aware reward modification for the cross-domain (cross-dynamics) RL setting~\cite{eysenbach2020off,liu2022dara}, which \textit{modifies the reward} by {using two classifiers} $q_\text{sas}(\cdot|\bs,\ba,\bs')$ and $q_\text{sa}(\cdot|\bs,\ba)$  that distinguish between the source-domain and target-domain data, \ie 
% \begin{align}
$r_\text{modified}(\bs, \ba, \bs') = r(\bs,\ba) + \log \frac{q_\text{sas}(\text{target}|\bs,\ba,\bs')}{q_\text{sas}(\text{source}|\bs,\ba,\bs')} - \log \frac{q_\text{sa}(\text{target}|\bs,\ba)}{q_\text{sa}(\text{source}|\bs,\ba)}$.
%  \label{eq:dara} 
% \end{align}
Compared to such a reward modification approach, our supported value optimization only learns a single transition model $\hat{T}_\text{target}$ that \textit{merely fits the target-domain data, while does not explicitly fit the source-domain data}. 
As we will show in our experiment, our support optimization enjoys more stable training and achieves better performance especially when the source domain involves complex and diverse data distributions.

\subsection{Practical Implementation}
\label{sec:implementation}

\begin{wrapfigure}{r}{0.50\textwidth}
\vspace{-22pt}
% \begin{figure}[H]
\centering
\begin{minipage}{0.50\textwidth}
\begin{algorithm}[H]
        \small 
	\caption{Training BOSA}
	\label{alg:bosa}   
	\begin{algorithmic}[1]
        \REQUIRE{Cross-domain data $\mathcal{D}_\text{mix} :=\mathcal{D}_\text{target} \cup \mathcal{D}_\text{source}$.}
		\STATE Initialize $\hat{\pi}_{\beta_\text{mix}}$, $\hat{T}_\text{target}$, $\pi_\theta$, and $Q_\phi$. \\
		{\textcolor{gray}{// Training $\hat{\pi}_{\beta_\text{mix}}$ and $\hat{T}_\text{target}$. }}  
		\STATE Fit the behavior policy $\hat{\pi}_{\beta_\text{mix}}$ with $\mathcal{D}_\text{mix}$.
		\STATE Fit the target transition $\hat{T}_\text{target}$ with $\mathcal{D}_\text{target}$. \\
		\textcolor{gray}{// Training $\pi_\theta$ and $Q_\phi$. } \\ 
		\FOR{$k=1,\cdots, K$} 
		\STATE Sample a batch of data from {$\mathcal{D}_\text{mix}$}.
		\STATE Learn $\pi_\theta$ with Equation 4. \textcolor{gray}{// Lagrangian} 
		\STATE Learn $Q_\phi$ with Equation 5. %\textcolor{gray}{/* (using an Ensemble)~*/} 
		\ENDFOR
	\end{algorithmic}
\end{algorithm}
\end{minipage}
% \end{figure}
\end{wrapfigure}
We now describe our instantiation of BOSA for supported policy and value optimization. 
First, instead of directly maximizing the log-likelihood objective for estimating $\hat{\pi}_{\beta_\text{mix}}$ and $\hat{T}_\text{target}$, we opt to use the conditional variational auto-encoder (CVAE~\cite{sohn2015learning}) for density estimation and likelihood inference (in Equations~\ref{eq:bosa-policy-support} and~\ref{eq:bosa-value-support}). 
Second, to tackle the constrained objective for supported policy optimization in Equation~\ref{eq:bosa-policy-support}, we optimize it by using a Lagrangian relaxation.  
Third, for the filter operator $\mathbbm{1}(\cdot)$ in supported value optimization (Equation~\ref{eq:bosa-value-support}), we learn an ensemble of target transition model $\hat{T}_\text{target}$ to maintain the model's uncertainty and take $1(\cdot > \epsilon)$ over the minimum of ensemble models. 
Empirically, we find the performance can be improved by increasing the ensemble size, but the improvement is saturated around 5. Thus, we learn 5 transition models in the ensemble. 
Finally, we summarize the proposed method for cross-domain offline RL in Algorithm~\ref{alg:bosa}. 
More details on the CVAE implementation, solving the Lagrangian relaxation, and hyper-parameters are provided in the appendix. 

\section{Experiments}
The goal of our empirical evaluation is to answer the following questions: 
\textbf{1)} Can BOSA improve offline data efficiency by leveraging the additional source-domain data and achieve better performance than prior alternative methods? 
\textbf{2)} In some cases there exists no additional source data pre-collected from other environments, can we retain the benefits of cross-domain data transfer by using BOSA? % over model-based RL and data augmentation? 
\textbf{3)} How do the different components of our method (hyper-parameters and regularization) influence BOSA's performance?

% , including the supported policy regularization (Equation~\ref{eq:bosa-policy-support}), the transition filtering (Equation~\ref{eq:bosa-value-support}), the value conservation (Equation~\ref{eq:bosa-value-support}), and the cross-domain data used for value optimization, 

In comparison, we consider the four most related offline RL methods: BCQ~\cite{fujimoto2019off}, CQL~\cite{kumar2020conservative}, MOPO~\cite{yu2020mopo}, SPOT~\cite{wu2022supported}, where BCQ motivates us the support constraints, CQL motivates us the conservation over policy optimization, MOPO is a representative model-based approach which enjoys data efficiency, and SPOT is the current state-of-the-art baseline which also performs supported policy optimization (corresponding to the objective in Equation~\ref{eq:bosa-policy-support}).

\begin{table*}[t]
	\centering
	\caption{Results on the single-domain and cross-domain offline RL. We take the baseline results (single-domain setting with 100\% D4RL) from their original papers. We average our results over 5 seeds and for each seed, we compute the normalized average score using 10 episodes. In the cross-domain setting, the numbers to the left of the arrow ($\to$) represent the scores trained on 100\% D4RL data, and the numbers to the right of that represent the scores trained on only 10\% D4RL data. In the left panel (single-domain setting), Average$^{\dagger}$ represents the average performance change when the offline data is reduced (100\%$\to$10\%). In the right panel (cross-domain setting), Average$^{\ddagger}$  represents the average performance difference between \textit{the cross-domain results} and \textit{the best results among baselines that are trained with 100\% D4RL}. In each line, we bold the best score among baselines that are trained with 10\% D4RL data, \ie including the single-domain 10\% D4RL setting and the cross-domain setting.  (ha: halfcheetah. ho: hopper. wa: walker2d. m: medium. mr: medium-replay. me: medium-expert.)} % ``Degradation$^{\ddagger}$'' denotes the number of tasks whose performance degrades due to the introduction of additional source domain data when only 10\% D4RL is available.  We can find that when we reduce of offline data, single-domain offline RL baselines suffer a large performance drop. 
	\vspace{-5pt}
	\begin{adjustbox}{max width=1.0\textwidth}
        \small 
	\begin{tabular}{cl
			S[table-format=2.1]@{\,\(\to \)\,}
			S[table-format=2.1]
			S[table-format=2.1]@{\,\(\to \)\,}
			S[table-format=2.1]
			S[table-format=2.1]@{\,\(\to \)\,}
			S[table-format=2.1]
			S[table-format=2.1]@{\,\(\to \)\,}
			S[table-format=2.1]
			S[table-format=2.1]S[table-format=2.1]S[table-format=2.1]S[table-format=2.1]
			S[table-format=2.1]@{\,\(\pm\)\,}
			l%S[table-format=2.1]
		}
		\toprule
		&       & \multicolumn{8}{c}{single-domain setting (100\% D4RL $\to$ 10\% D4RL)}        & \multicolumn{6}{c}{cross-domain setting (10\% D4RL + source data)} \\
		\cmidrule(lr){3-10} \cmidrule(lr){11-16}
		&       & \multicolumn{2}{c}{BCQ} & \multicolumn{2}{c}{MOPO} & \multicolumn{2}{c}{CQL} & \multicolumn{2}{c}{SPOT} & \multicolumn{1}{l}{BCQ} & \multicolumn{1}{l}{MOPO} & \multicolumn{1}{l}{CQL} & \multicolumn{1}{l}{SPOT} & \multicolumn{2}{c}{BOSA} \\
		\midrule
		\multirow{9}[1]{*}{mass} & ha-m  & 40.7  & 37.6  & 42.3  & 3.2  & 44.4  & 35.4  & 58.4  & 45.4   & 35.1  & 6.4   & 32.2  & 50.3  & \textbf{58.3}  & 2.5 \\
		& ha-mr & 38.2  & 1.1   & 53.1  & -0.1  & 46.2  & 0.6   & 52.2  & 9.8   & \multicolumn{1}{c}{\textbf{40.1}\ \ }  & 10.2  & 3.3    & 37.6  & 37.2 & 0.7  \\
		& ha-me & 64.7  & 37.3  & 63.5  & 4.2   & 62.4  & -3.3  & 86.9  & 46.2  & 26.4  & 8.9   & 12.9   & 33.8  & \textbf{51.6}  & 0.1 \\
		& ho-m  & 54.5  & 37.1  & 28    & 4.1   & 58    & 43    & 86    & 62.5  & 25.7  & 5     & 44.9  &  \textbf{85.9} & 82.4  & 2.1 \\
		& ho-mr & 33.1  & 9.3   & 67.5  & 1     & 48.6  & 9.6   & 100.2 & 13.7  & 28.7  & 5.5   & 1.4   & 15.5  & \textbf{39.7}    & 0.1 \\
		& ho-me & 110.9 & 58    & 23.7  & 1.6   & 98.7  & 59.7  & 99.3  & 69    & 75.4  & 4.8   & 53.6  & 75.5  & \textbf{104.2} & 0.5 \\
		& wa-m  & 53.1  & 32.8  & 17.8  & 7     & 79.2  & 42.9  & 86.4  & 65.4  & 50.9  & 5.7   & 80    & 22.5  & \textbf{83}    & 2.9 \\
		& wa-mr & 15    & 6.9   & 39    & 5.1   & 26.7  & 4.6   & 91.6  & 18.6  & 14.9  & 3.1   & 0.8   & 16    & \textbf{21.4}  & 2 \\
		& wa-me & 57.5  & 32.5  & 44.6  & 5.3   & 111   & 49.5  & 112   & 84    & 55.2  & 5.5   & 63.5  & 14.3  & \textbf{86.5}  & 0.6 \\
		\midrule
		\multirow{9}[2]{*}{joints} & ha-m  & 40.7  & 37.6  & 42.3  & 3.2   & 44.4  & 35.4  & 58.4  & 45.4  & 40    & 3.5   & 40.7  & 50.1    & \textbf{56.2}  & 0.27 \\
		& ha-mr & 38.2  & 1.1   & 53.1  & -0.1  & 46.2  & 0.6   & 52.2  & 9.8   & 39.4  & 2.6   & 2     & 41 & \textbf{51.3}& 1.1 \\
		& ha-me & 64.7  & 37.3  & 63.5  & 4.2   & 62.4  & -3.3  & 86.9  & 46.2  & \multicolumn{1}{c}{\textbf{55.3}\ \ }  & 1.5   & 7.7   & 38 .1   & 52.8  & 0.45 \\
		& ho-m  & 54.5  & 37.1  & 28    & 4.1   & 58    & 43    & 86    & 62.5  & 49    & 9.2   & 58    & 41.5  & \textbf{78}    & 7.3 \\
		& ho-mr & 33.1  & 9.3   & 67.5  & 1     & 48.6  & 9.6   & 100.2 & 13.7  & 23.8  & 2.3   & 2.6   & 23    & \textbf{32.7}    & 1.3 \\
		& ho-me & 110.9 & 58    & 23.7  & 1.6   & 98.7  & 59.7  & 99.3  & 69    & 96    & 6.1   & 73.4  & 52    & \textbf{96.4}  & 0.5 \\
		& wa-m  & 53.1  & 32.8  & 17.8  & 7     & 79.2  & 42.9  & 86.4  & 65.4  & 44.9  & 7.8   & 73.2  & 38.8  & \textbf{86.5}  & 5.6 \\
		& wa-mr & 15    & 6.9   & 39    & 5.1   & 26.7  & 4.6   & 91.6  & 18.6  & 9.8   & 9.3   & 1.4   & 10.7  & \textbf{38.2}  & 4.7 \\
		& wa-me & 57.5  & 32.5  & 44.6  & 5.3   & 111   & 49.5  & 112   & 84    & 40.6  & 15.2  & \multicolumn{1}{c}{\textbf{109.9}\ \ } & 74.3  & 85.8  & 0.3 \\
		\midrule
%		\multicolumn{2}{l}{Average$^{\dagger}$ ($\%$)} & 
%		\multicolumn{2}{r}{$-48.3\%$}  &  \multicolumn{2}{r}{$-88.7\%$} & \multicolumn{2}{r}{$-61.5\%$} & \multicolumn{2}{r}{$-47.1\%$} & 
%		\multicolumn{1}{r}{$-50.1\%$} & \multicolumn{1}{r}{$-92.5\%$} & \multicolumn{1}{r}{$-59.4\%$} & \multicolumn{1}{r}{$-50.1\%$} & \multicolumn{2}{c}{{$-$}\textbf{31.2}{$\%$}} \\
		\multicolumn{2}{l}{Average$^{\dagger}$ ($\%$)} & 
		\multicolumn{2}{r}{$-48.3\%$}  &  \multicolumn{2}{r}{$-88.7\%$} & \multicolumn{2}{r}{$-61.5\%$} & \multicolumn{2}{r}{$-47.1\%$} & \\
		\multicolumn{2}{l}{Average$^{\ddagger}$ ($\%$)} & 
		\multicolumn{8}{r}{ }  &  
		\multicolumn{1}{r}{$-50.1\%$} & \multicolumn{1}{r}{$-92.5\%$} & \multicolumn{1}{r}{$-59.4\%$} & \multicolumn{1}{r}{$-50.9\%$} & \multicolumn{2}{c}{{$-$}\textbf{25.6}{$\%$}} \\
%		\multicolumn{2}{l}{Degradation$^{\ddagger}$} & \multicolumn{2}{r}{ }  &  \multicolumn{2}{r}{ } & \multicolumn{2}{r}{ } & \multicolumn{2}{r}{ } & {0/18} & {0/18} & {0/18} & {0/18} & \multicolumn{2}{c}{0/18} \\
		\bottomrule
	\end{tabular}
	\end{adjustbox}
	\label{tab:exp-single-vs-cross}
\end{table*}

\begin{table}[t]
	\centering
	\caption{Comparison on cross-domain tasks, where \textit{DARA* denotes the best score} among the offline RL baselines (BCQ, CQL, and MOPO) when using dynamics-aware reward modification (DARA).}
	\vspace{1pt}
        \small 
	\begin{adjustbox}{max width=0.975\columnwidth}
		\begin{tabular}{llS
				[table-format=2.1]S[table-format=2.1]
    			S[table-format=2.1]@{\,\(\pm\)\,}
				l|
                    llS
				[table-format=2.1]S[table-format=2.1]
    			S[table-format=2.1]@{\,\(\pm\)\,}
				l
                    }
			\toprule
			\multicolumn{1}{l}{mass} & \multicolumn{1}{l}{Finetune} & \multicolumn{1}{l}{DARA*} & \multicolumn{1}{l}{MABE} & \multicolumn{2}{c}{BOSA}&\multicolumn{1}{c}{joints} & \multicolumn{1}{l}{Finetune} & \multicolumn{1}{l}{DARA*} & \multicolumn{1}{l}{MABE} & \multicolumn{2}{c}{BOSA} \\
			\midrule
			  ho-m  & 44.5  & 59.3  & 23.1  & \textbf{80.5}  & 1.7 & ho-m  & 52.5  & 58    & 57.7  & \textbf{78}    & 7.3  \\
			  ho-mr & 27.5  & {34.1}  & 20.4  & \textbf{39.7}   & 0.1 & ho-mr & 29.6  & 32 & \multicolumn{1}{c}{\textbf{35.4}\ \ }  & 32.7    & 1.3 \\
			ho-me & 85.9  & 99.7  & 38.9  & \textbf{104.2} & 0.5 &ho-me & 107.3 & \multicolumn{1}{c}{\textbf{109}\ \ \ \ }   & 104.8 & 96.4  & 0.5 \\
			wa-m  & 72.3  & 81.7  & 56.7  & \textbf{83}    & 2.9 & wa-m  & 76.6  & 81.2  & 48.7  & \textbf{86.5}  & 5.6\\
			  wa-mr & 10.4  & 15.1  & 12.5  & \textbf{21.4}  & 2& wa-mr & 13.5  & 16.4  & 1.6   & \textbf{38.2}  & 4.7 \\
			  wa-me & 68.6  & \multicolumn{1}{c}{\textbf{93.3}\ \ }  & 82.7  & 86.5  & 0.6& wa-me & 104   & \multicolumn{1}{c}{\textbf{116.5}\ \ } & 82.6  & 85.8  & 0.3 \\
			\midrule
			\multicolumn{1}{l}{Sum} & 309.2 & 383.2 & 234.3 & \multicolumn{1}{l}{\textbf{415.3}}&	&		\multicolumn{1}{l}{Sum} & 383.5 & 413.1 & 330.8 & \multicolumn{1}{l}{\textbf{417.6}} \\
			\bottomrule
		\end{tabular}%
	\end{adjustbox}
	\label{tab:cross-dara}%
	\vspace{-5pt}
\end{table}%

\vspace{-3pt}
\paragraph{Offline cross-domain data transfer.} 
To answer the first question, we use the D4RL~\cite{d4rl2020Fu} offline data as the target domain and use the similar cross-domain dynamics modification utilized in DARA~\cite{liu2022dara} to collect source-domain data. Specifically, the source-domain data are collected by modifying the body \textbf{mass} or adding noise to \textbf{joints} of the agent and then following the same data-collection procedure as in D4RL. 
%We note that the performance of current advanced offline methods tends to be saturated in D4RL. 
To study the data efficiency and make the cross-domain setting tractable, we only use 10\% of the D4RL data in our target domain.

In Table~\ref{tab:exp-single-vs-cross}, we provide the results of different methods using single-domain offline data or cross-domain data. We can see that in the single-domain setting, BCQ, CQL, MOPO, and SPOT both suffer a large performance drop when we reduce the training data size from 100\% D4RL to 10\% D4RL. Further, using additional source-domain data (\ie the cross-domain setting) also does not provide a clear performance improvement compared to that using only the target-domain data (10\% D4RL). We can observe that in 6 out of the 18 tasks, cross-domain data even brings performance degradation for CQL. The main reason is that although cross-domain setting introduces additional source-domain data, it also raises the challenge of ODD transition dynamics. Aiming at improving the data efficiency and eliminating ODD transition dynamics, we can observe that our BOSA brings significant performance improvement (in 14 out of 18 tasks) compared to baselines when using 10\% D4RL data. 
In comparison to the best performance of baselines when using 100\% D4RL data, BOSA receives the fewest performance degradation ($-25.6\%$) among baselines when using 10\% D4RL. 
% which even outperforms the best performance of baselines when using 100\% D4RL data in \textcolor{red}{five} out of 18 tasks. 

Then, we compare BOSA to cross-domain offline RL baselines: DARA~\cite{liu2022dara} and MABE~\cite{cang2021behavioral}, where DARA conducts the dynamics-aware reward modification and MABE learns a cross-domain behavior prior. Additionally, we also introduce a fine-tuning baseline (Finetune), which first trains a policy on the source-domain data and then finetunes it over the target-domain data (10\% D4RL). 
We show the results in Table~\ref{tab:cross-dara}. We can see that BOSA is competitive with DARA (reward modification) in 9 out of 12 tasks and outperforms or matches Finetune and MABE on all 12 cross-domain tasks. 

% 5/5 target: 50% medium-expert| source: 100% medium
%In addition, in order to more intuitively verify the ability of BOSA to extract the available state-action pairs from source dataset. We selected 50\% of the medium-expert datasets in the three tasks of hopper, halfcheetah and walker2d as the target domain, and selected the corresponding medium dataset of the three tasks as the source domain, then perform BOSA on those three tasks, and results are shown in Table. . We can see that, BOSA outperformed or equalled SPOT trained with 100\% medium-expert data on all three tasks, further displaying that BOSA can indeed extract helpful state-action pairs from other datasets.

%Decision Transformer achieves the highest scores in a
%majority of the tasks and is competitive with the state of the art in the remaining tasks.
%
%We observe that MABE either matches or outperforms prior methods
%in a majority of the tasks, and achieves the highest average score.\

\begin{table}[t]
	\centering
	\caption{When source-domain data is not available, we can use a (sub-optimal) pseudo-transition model or data augmentation to generate new transitions (source-domain data). We can find that naively augmenting (target) offline data will inevitably result in negative performance improvement while our BOSA can eliminate the transition dynamics mismatch and contribute to gained performance in 11 out of 12 tasks. (\textit{aug.}: augmentation)}
    \small
	\begin{tabular}{ll
			S[table-format=2.1]
			S[table-format=2.1]@{\,\(\pm\)\,}
			l%S[table-format=1.1]
			S[table-format=2.1]@{\,\(\pm\)\,}
			l@{\,\( \ \)\,}|l%S[table-format=1.1]
                S[table-format=2.1]
			S[table-format=2.1]@{\,\(\pm\)\,}
			l%S[table-format=1.1]
			S[table-format=2.1]@{\,\(\pm\)\,}
			l@{\,\( \ \)\,}l%S[table-format=1.1]
		} %\multicolumn{1}{c}{SPOT}
		\toprule
		&       &  \multicolumn{1}{c}{\multirow{2}[1]{*}{\makecell[c]{SPOT \\ (10\%)}}}  & \multicolumn{4}{c}{cross-domain}&       &  \multicolumn{1}{c}{\multirow{2}[1]{*}{\makecell[c]{SPOT \\ (10\%)}}}  & \multicolumn{4}{c}{cross-domain} \\
		\cmidrule(lr){4-7}
            \cmidrule(lr){10-13}
		&  &  & \multicolumn{2}{c}{SPOT} & \multicolumn{2}{l}{BOSA} &  &  & \multicolumn{2}{c}{SPOT} & \multicolumn{2}{l}{BOSA}\\
		\cmidrule(lr){2-13}
		  ho-m &\multirow{6}[2]{*}{\makecell[c]{pseudo-\\model \\ \textit{aug.}}}  & 62.5  & 0.7  & 0.6   & \textbf{66.6}  & 1.6 &\multirow{6}[2]{*}{\makecell[c]{noise \\ \textit{aug.}}}  & 62.5  & 29    & 5     & \textbf{76.6}  & 9.2  \\
		 ho-mr& & 13.7  & 0.6   & 0.3   & \textbf{14.2}   & 0.2 &  & 13.7  & 12.8  & 1.5   & \textbf{16.5}  & 0.3  \\
		 ho-me& & 69    & 0.6   & 0.3   & \textbf{70.7}   & 2 &  & 69    & 66.3  & 1.9   & \textbf{78.5}  & 1.7  \\
	   wa-m&  & 65.4  & -2.1  & 1.9   & \textbf{76.7}  & 0.7 && 65.4  & 67.4  & 10    & \textbf{78.6}  & 3.3   \\
		wa-mr&  & 18.6  & 1.4   & 0.2   & \textbf{20.1}  & 2.3 & & \multicolumn{1}{c}{\textbf{18.6}\ \ }  & 14.8  & 2.2   & 12.6  & 1  \\
		wa-me&  & 84   & 0.5   & 1.3  & \textbf{84.8}  & 0.4 &   & 84    & 82.6  & 0.1   & \textbf{84}    & 0.8 \\
		\bottomrule
	\end{tabular}%
	\label{tab:exp-data-augmentation}%
\end{table}%

\vspace{-3pt}
\paragraph{Model-based RL and (noising) data augmentation.}  
For the second question, we answer it affirmatively. If we can not access additional source-domain data, we can learn a sub-optimal pseudo-transition model and use the learned model to generate new cross-domain transitions. Alternatively, we can also employ data augmentation (adding noise) to generate new cross-domain transitions. Then, we can directly treat the generated transitions as the source data. %we first use the (limited) target domain data to learn a pseudo-transition model. Then, we can use the learned pseudo-transition model or directly employ data augmentation techniques to generate new transitions and treat the generated transitions as the source domain. 
More importantly, here we do not require the learned pseudo-transition model to be optimal (in model-based setting) or to delicately balance the amplitude of noises (in data augmentation). Although the generated source-domain data might involve OOD transitions, BOSA can filter out mismatched transitions, and preserve the target-relevant and beneficial transitions when performing supported policy and value optimization. 

We provide the comparison results in Table~\ref{tab:exp-data-augmentation}. 
We can find that naively using a (sub-optimal) pseudo-transition model or employing data augmentation does not improve or even hurts their performance in most tasks. In contrast, BOSA can improve the cross-domain performance in 11 out of 12 tasks, thus facilitating effective cross-domain offline RL by automatically generating the source-domain data.

\begin{figure*}[ht]
	% \vspace{-2pt}
	\begin{center}
    \includegraphics[scale=0.21]{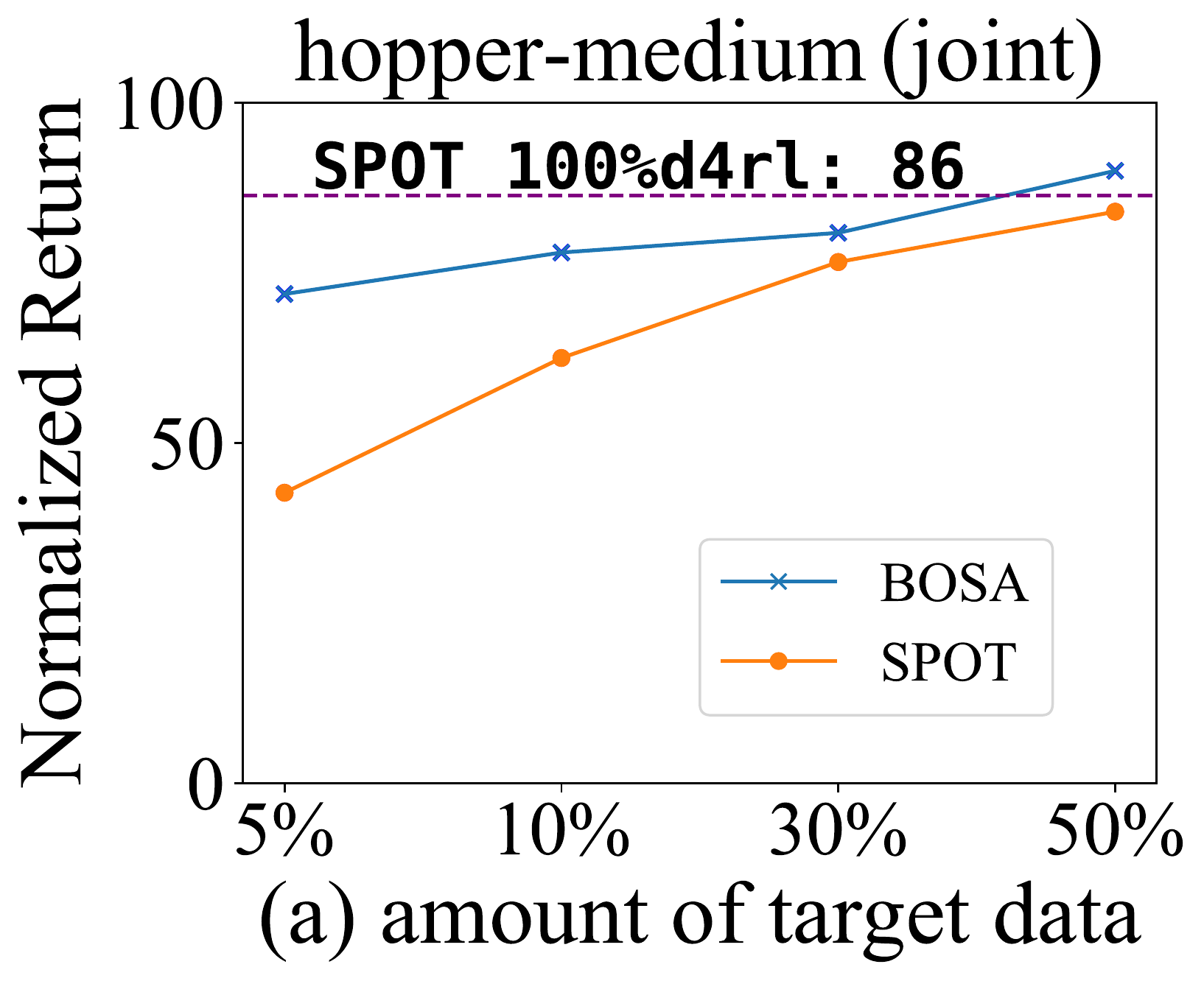}	
    \includegraphics[scale=0.21]{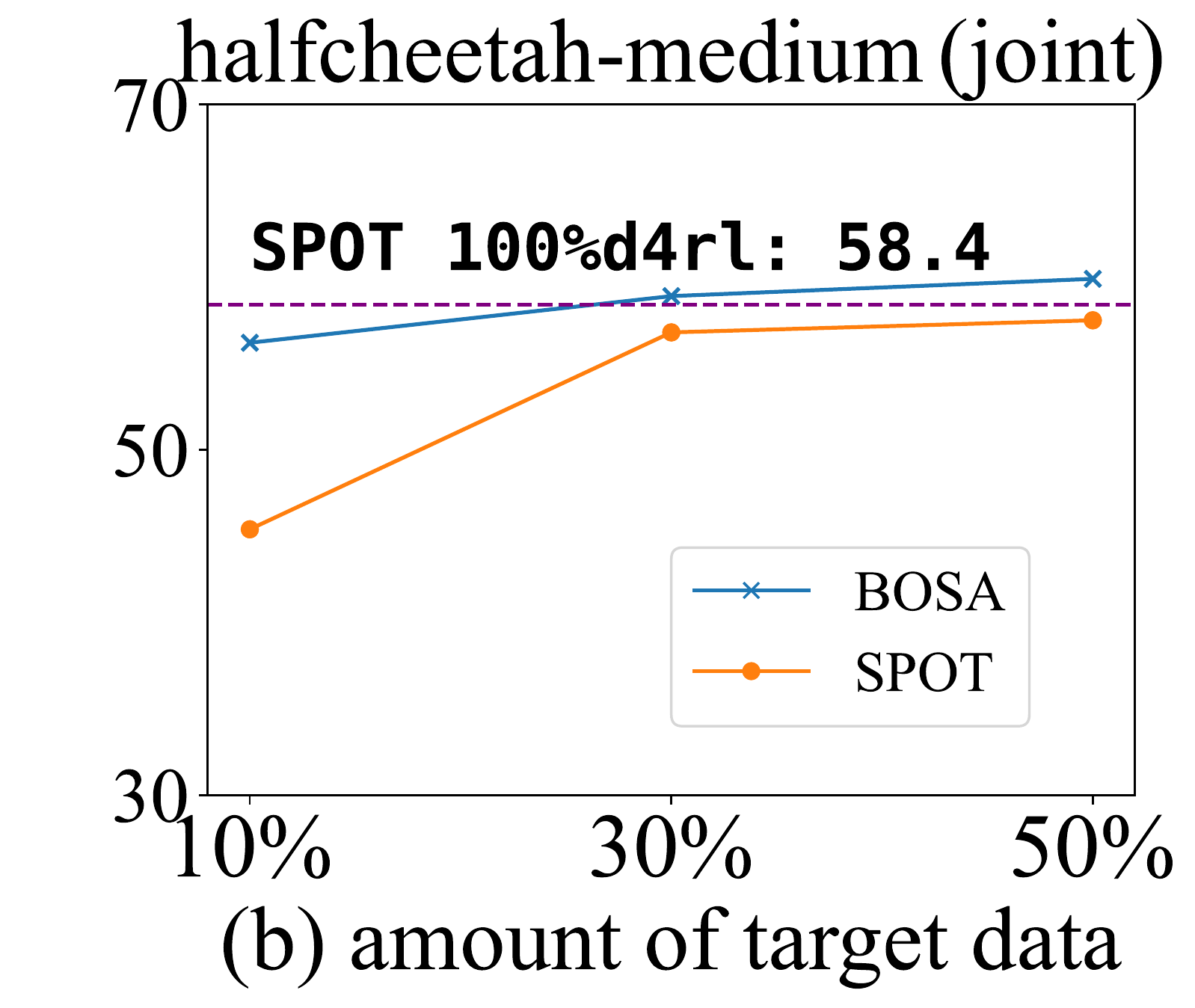}
    \includegraphics[scale=0.21]{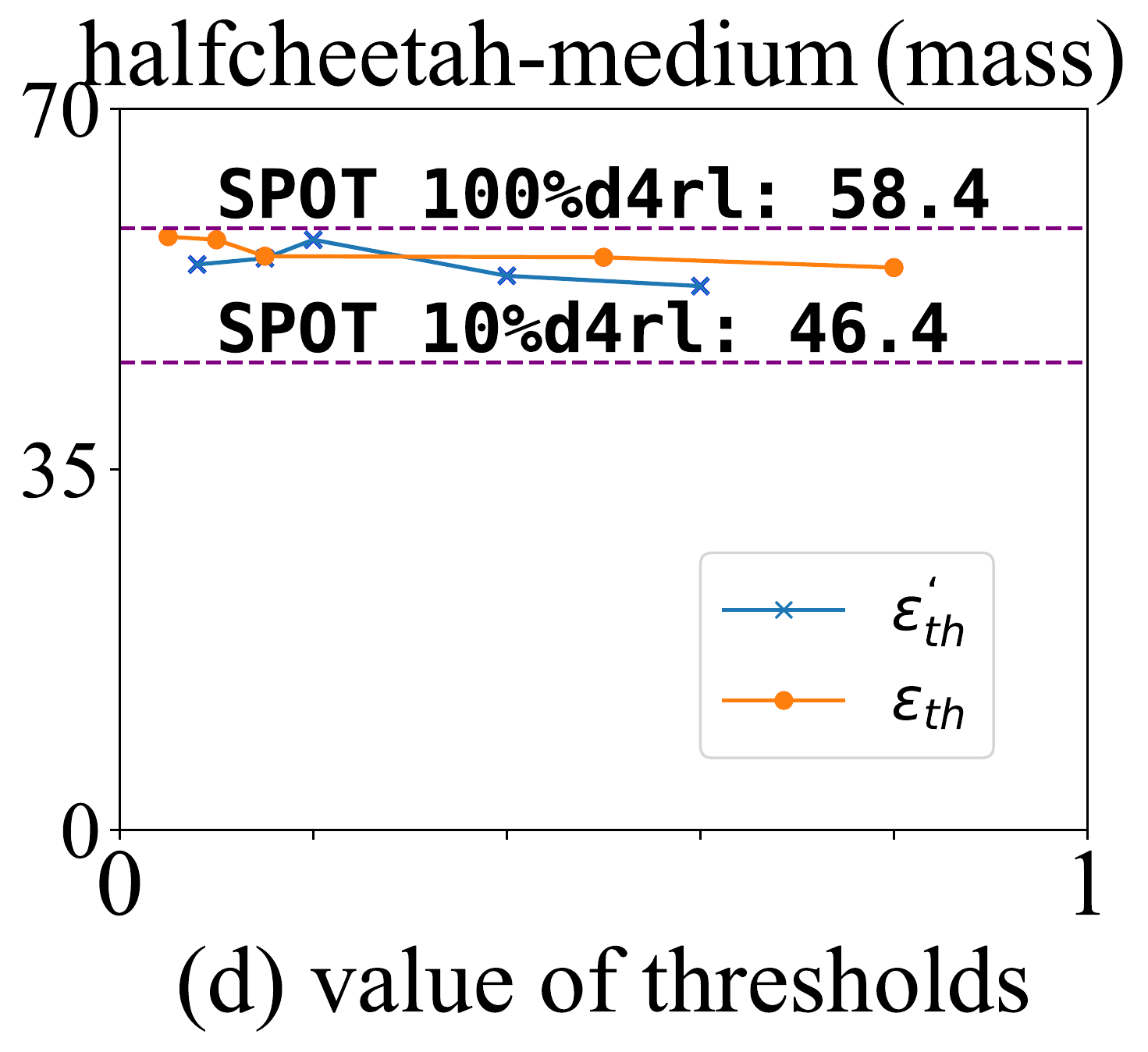}
  	\includegraphics[scale=0.21]{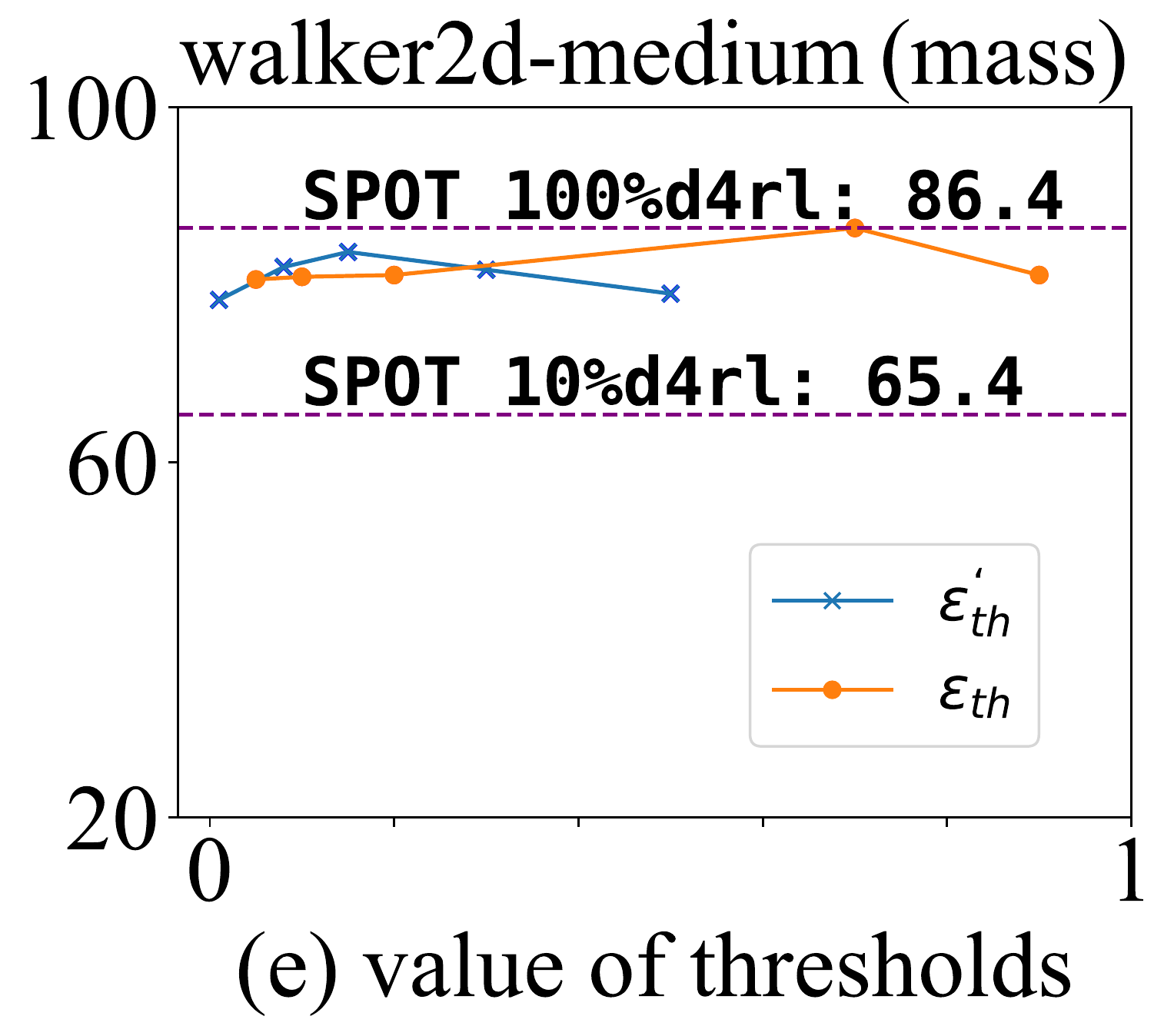}
	\end{center}
	\vspace{-5pt}
	\caption{Sensitivity on the amount of target-domain data (a, b) and the thresholds $\epsilon_{\text{th}}$ and $\epsilon_{\text{th}}'$ (c, d). Across a range of amount of target data and thresholds, BOSA consistently achieves better results compared to SPOT.}
	\label{fig:ablation-num-thresholds}
	\vspace{-12pt}
\end{figure*}

\vspace{-3pt}
\paragraph{Ablation study.}  
To answer the third question, we conduct a thorough ablation study on BOSA. 
We first investigate the sensitivity of BOSA on the amount of target-domain data. In Figure~\ref{fig:ablation-num-thresholds} (a, b), we present the cross-domain results across 5\%, 10\%, 30\%, and 50\% of target data. The results show that increasing the amount of target data generally improves the performance of both SPOT and BOSA, showing that data amount is critical to the offline performance. 
Consistently, BOSA achieves better performance than SPOT across a wide range of target data, showing that BOSA can contribute to the target-domain sample efficiency by using additional cross-domain data. 

Then, we study the hyper-parameter sensitivity on the thresholds $\epsilon_{\text{th}}$ and $\epsilon_{\text{th}}'$ in supported policy and value optimization respectively. We provide the sensitivity results in Figure~\ref{fig:ablation-num-thresholds} (c, d).

\begin{figure*}[t]
	%	\vspace{-2pt}
	\begin{center}
  \includegraphics[scale=0.23]{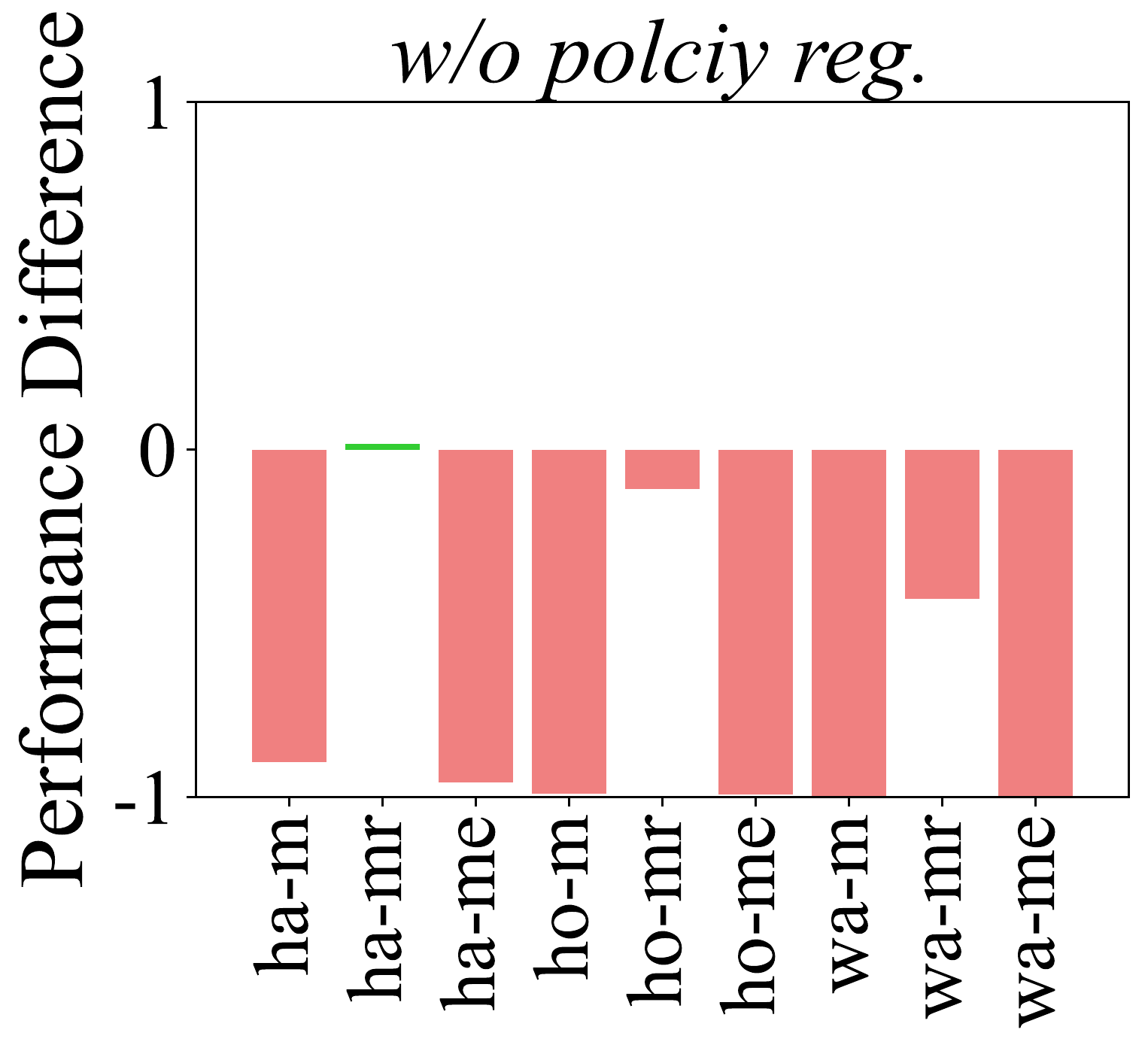}
		\includegraphics[scale=0.23]{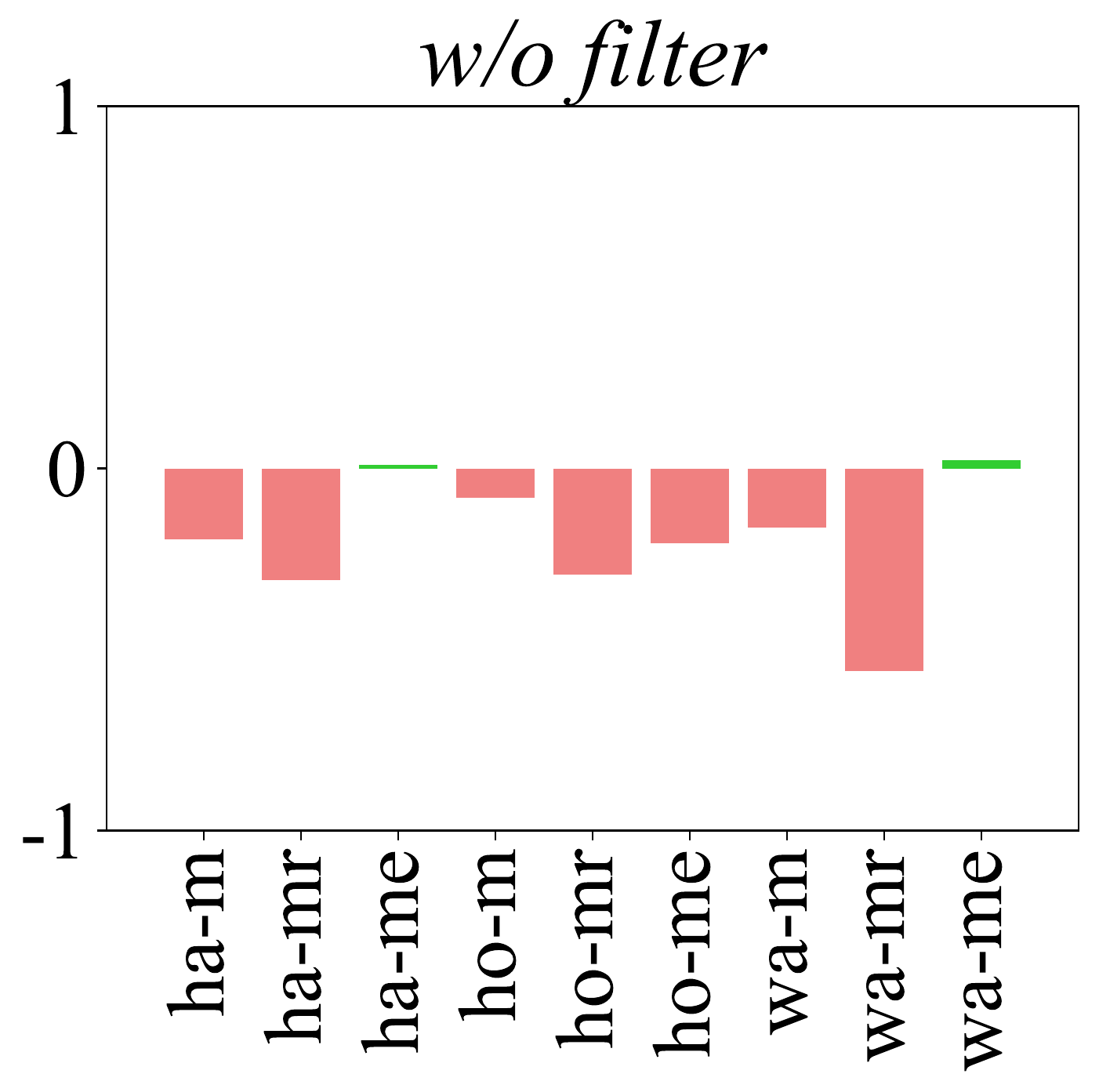}
		\includegraphics[scale=0.23]{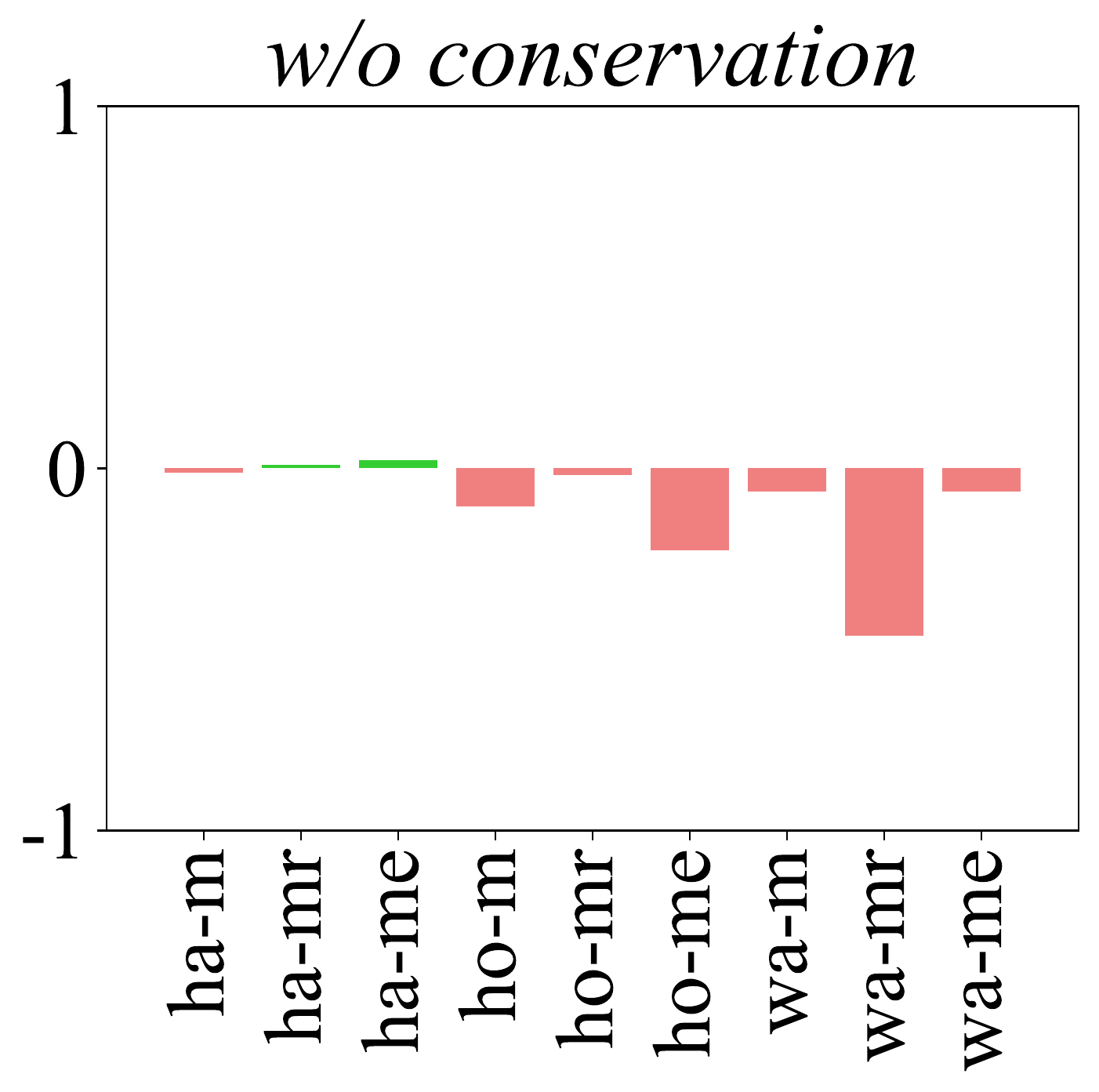}
		\includegraphics[scale=0.23]{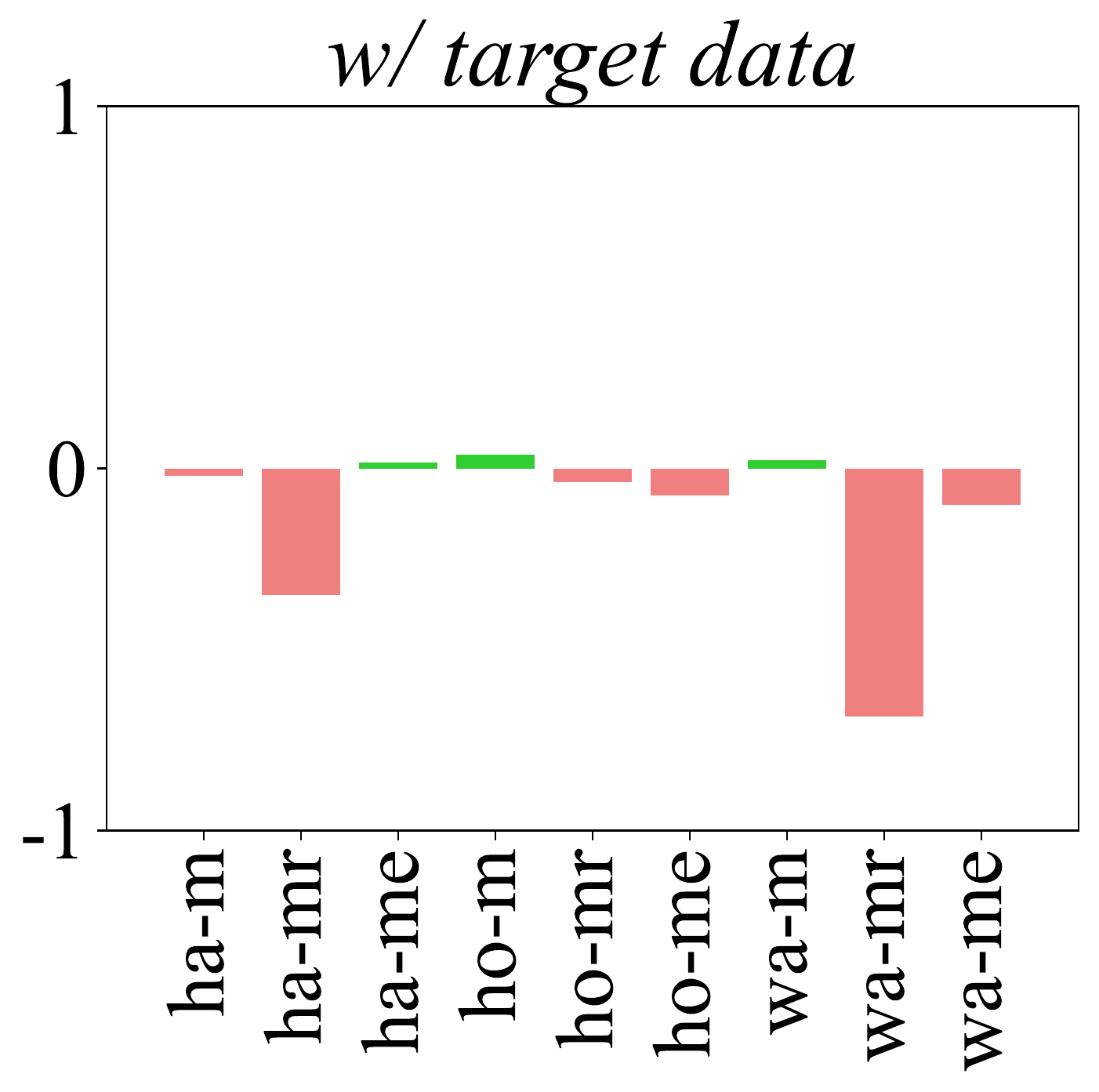}
	\end{center}
	\vspace{-5pt}
\caption{Performance changes when we ablate different components of our method. The y-axis denotes the percentage improvement of the ablated BOSA versus the full BOSA implementation. We can find that across a range of tasks, full BOSA preserves the best performance.}
	\label{fig:ablation-vs}
	\vspace{-12pt}
\end{figure*}

To understand how the choice of different components of BOSA affects its performance, we continue conducting the following ablations: 
\textbf{1)}~\textit{w/o policy reg.}: we remove the supported policy regularization in Equation~\ref{eq:bosa-policy-support}. % to check whether the support constraints (filter operator) in value optimization can accommodate the potential OOD state actions. 
\textbf{2)}~\textit{w/o filter}: we remove the filter operator in value optimization (Equation~\ref{eq:bosa-value-support}). % to investigate the cross-domain performance without explicit transition adaptation. 
\textbf{3)}~\textit{w/o conservation}: we remove the conservation regularization (Equation~\ref{eq:bosa-value-support}) in value optimization. % to check how the usage of conservation affects the performance. 
\textbf{4)}~\textit{w/ target data}: To eliminate the potential OOD transition dynamics, one can directly perform value optimization over the target-domain data. Thus, we replace the expectation of Equation~\ref{eq:bosa-value-support} by target-domain data $\mathcal{D}_\text{target}$ and remove the corresponding filter operator. 
Due to page length limitations, we leave the detailed implementation w.r.t. these ablation experiments to the appendix. %supplementary material. 

The results of our ablation studies are shown in Figure~\ref{fig:ablation-vs}, where we present the percent difference in performance when removing the corresponding components of BOSA. 
As expected, when we remove the policy regularization (\textit{w/o policy reg.}), the performance of BOSA degrades dramatically, showing that OOD state action is still the dominant issue in cross-domain offline RL. Further, removing the filter operator (\textit{w/o filter}) also causes a substantial performance drop of around 20\%. In comparison, \textit{w/o policy reg.} and \textit{w/ target data} suffer less performance degradation, while still causing unstable training and a large variance across a range of tasks.

\vspace{-1pt}
\section{Conclusion}

In this paper, we formalize the cross-domain offline RL in an effort to improve offline data efficiency. 
Beyond the common OOD state actions issue,% in naive (single-domain) offline RL, 
we identify a new challenge of OOD transition dynamics in the cross-domain offline setting. To address this problem, we propose supported policy and value optimization, which explicitly regularizes the policy and value optimization with in-support transitions. 
% In the cross-domain setting, our method (BOSA) can leverage the source-domain data and enable data-efficient learning, while eliminating the potential OOD transition dynamics. 
Empirically, we demonstrate in a variety of offline cross-domain tasks, BOSA can outperform existing cross-domain baselines. 
Further, we also show that when the assumed source-domain data is not accessible, BOSA can be naturally plugged into model-based RL and (noising) data augmentation techniques, and enjoys broad flexibility for cross-domain offline data transfer.

\begin{wrapfigure}{r}{0.49\textwidth}
	\vspace{-15pt}
	\begin{center}
 % domain_show.png
        \includegraphics[scale=0.22]{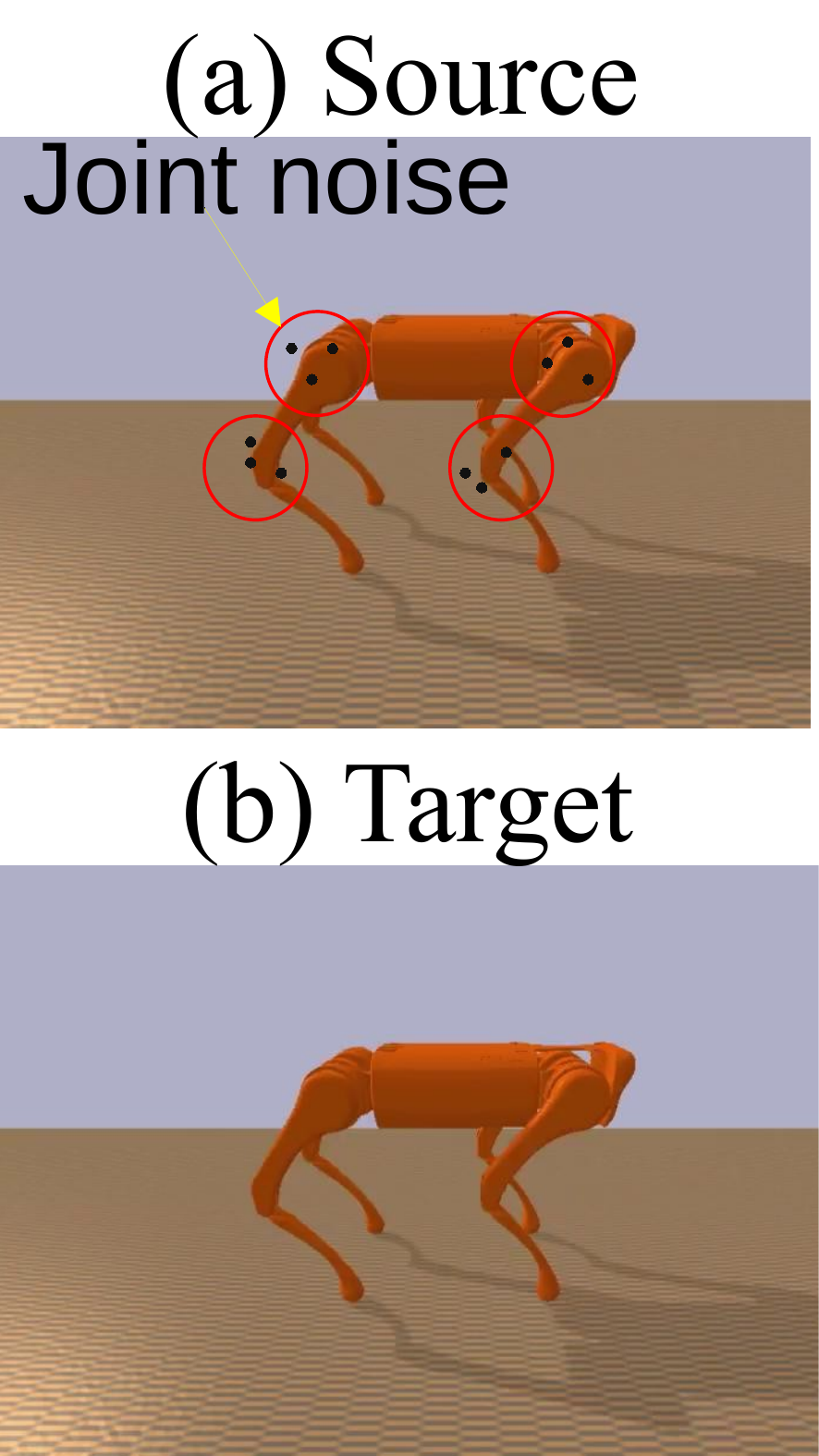}
\includegraphics[scale=0.30]{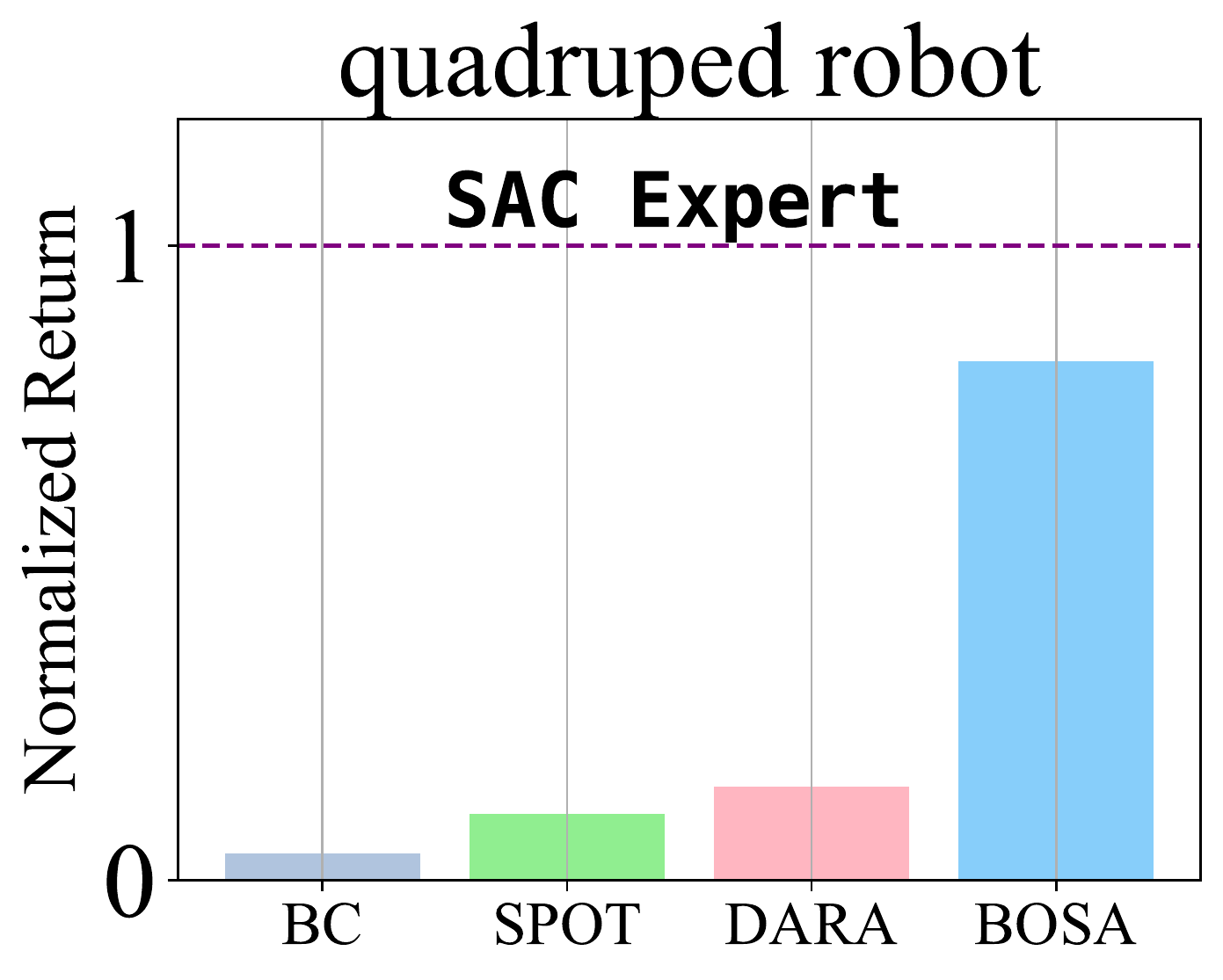}

	\end{center}
	\vspace{-2pt}
	\caption{Cross-domain offline RL results on a simulated cross-domain quadruped locomotion task, where the y-axis denotes the normalized return recorded on the target domain. }
        \label{fig:bosa-quadruped}
	\vspace{-10pt}
\end{wrapfigure}
\vspace{-3pt}
\paragraph{Limitations and future work.} 

Aiming at improving the offline RL sample efficiency, we expect to deploy BOSA to more complex quadruped locomotion tasks. To do so, we have validated BOSA on two simulated quadruped locomotion environments (with transition dynamics mismatch, see details in appendix) and, following D4RL, collected $2\times10^6$ (source-domain) and $3\times10^4$ (target-domain) medium-expert offline transitions. Then, we conduct the standard offline RL training paradigm and test the performance in the target simulation environment. In Figure~\ref{fig:bosa-quadruped}, we provide the comparison results of BC, SPOT, DARA, and BOSA. We can find that BOSA achieves significant performance gains in this cross-domain task, showing a greater potential that awaits future real-world quadruped robots.

Regarding the methodological assumption made in this paper, we assume source and target domains share the same state space, action space, and reward function. In more general real-world settings, such the state/action space and reward may be also different between the two domains. Thus,  future works could further exploit source domain data with different state/action spaces, perhaps with different reward functions, to further improve the data efficiency of offline RL methods in low-data regimes. 
Another interesting direction is to use support policy/value constraints for online RL and safe RL~\cite{schulman2015trust,liu2022constrained,gu2022review}. Similar to those trust region RL methods, we can use support constraints (instead of distribution matching objectives) to constrain the learning policy.

% \section*{References}

% \bibliographystyle{unsrtnat}

\bibliographystyle{plainnat}
\bibliography{ijcai23.bib}

\newpage

\section{Social Impacts}

Typically, offline RL holds the promise of enabling RL agents to learn complex behaviors through fixed and static offline data. However, realizing this promise for data-limited tasks in the real world requires mechanisms to improve offline learning efficiency. 
This paper explicitly proposes cross-domain transfer and data augmentation for such data-limited offline scenes.  We believe our work is an important step towards data-efficient offline RL while offering significant improvement, improving offline RL transferability, and providing a promising approach for real-world offline RL participation.

\section{Implementation Details}

\subsection{CVAE Implementation}

To estimate the empirical behavior policy $\hat{\pi}_{\beta_\text{mix}}(\ba|\bs)$ and the target-domain transition dynamics $\hat{T}_\text{target}(\bs'|\bs,\ba)$, we use conditional variational auto-encoder (CVAE)~\cite{sohn2015learning} to maximize a lower bound of the log-likelihood objective. 

Specifically, we can write the variational lower bound of a model $\log p_\theta(\by|\bx)$ as follows:
\begin{align}
\log p_\theta(\by|\bx) &\geq - \text{KL}(q_\phi(\bz|\bx,\by) || p_\theta(\bz|\bx)) \nonumber + \mathbb{E}_{q_\phi(\bz|\bx,\by)}\left[ p_\theta(\by|\bx,\bz) \right],
\end{align}
and the corresponding empirical lower bound as:
\begin{align}
\hat{\mathcal{L}}_\text{CVAE}  &= -\text{KL}(q_\phi(\bz|\bx,\by) || p_\theta(\bz|\bx)) \nonumber + \frac{1}{L}\sum_{l=1}^{L} \log p_\theta(\by|\bx,\bz^l), \nonumber
\end{align}
where $\text{KL}$ denotes the KL divergence and $\bz^{l} \sim q_\phi(\bz|\bx,\by)$ denotes the $l$-th sample of the encoder network $q_\phi$ (using the reparameterization trick). 

\textbf{Training $\hat{\pi}_{\beta_\text{mix}}(\ba|\bs)$.} We thus estimate the empirical behavior policy $\hat{\pi}_{\beta_\text{mix}}$ by maximizing
\begin{align}
\hat{\mathcal{L}}_{\text{CVAE}(\hat{\pi}_{\beta_\text{mix}})}  &= -\text{KL}(q_\phi(\bz|\bs,\ba) || p_\theta(\bz|\bs)) \nonumber + \frac{1}{L}\sum_{l=1}^{L} \log p_\theta(\ba|\bs,\bz^l). \nonumber
\end{align}

For simplicity, here we abuse the notations $\phi$ and $\theta$ to represent the parameters of the encoder and decoder networks for optimizing $\hat{\pi}_{\beta_\text{mix}}(\ba|\bs)$ and $\hat{T}_\text{target}(\bs'|\bs,\ba)$. 

\textbf{Inference.} Once the model parameters (encoder $\phi$ and decoder $\theta$) are learned, we can use the importance sampling to estimate the conditional likelihood. Taking the model $p_\theta(\bx|\by)$ as an example, we estimate the likelihood with:
\begin{align}
p_\theta(\by|\bx) \approx \frac{1}{L} \sum_{l=1}^{L}\frac{p_\theta(\by|\bx,\bz^l )p_\theta(\bz^l | \bx)}{q_\phi(\bz^l | \bx,\by )},\nonumber
\end{align}
where $\bz^l \sim q_\phi(\bz|\bx,\by)$.

\subsection{Lagrangian Relaxation}
To optimize the constrained policy optimization objective,
\begin{align}
\max_{\pi_\theta} \ \mathcal{J}_{\mathcal{D}_\text{mix}}(\pi_\theta) &:= \mathbb{E}_{\bs \sim \mathcal{D}_\text{mix}, \ba\sim\pi_\theta(\ba|\bs)}\left[ Q_\phi(\bs, \ba) \right], \nonumber
\text{s.t.} \ \mathbb{E}_{\bs\sim\mathcal{D}_\text{mix}}\left[ \log\hat{\pi}_{\beta_\text{mix}}(\pi_\theta(\bs)|\bs) \right] > \epsilon_\text{th}, \nonumber
\end{align} 

in Equation 4 in the main paper, we construct a Lagrangian  relaxation: 
\begin{align}
\label{eq:a_filtering}
\max_{\pi_\theta} \min_{\lambda > 0} \ &\mathbb{E}_{\bs \sim \mathcal{D}_\text{mix}, \ba\sim\pi_\theta(\ba|\bs)}\left[ Q_\phi(\bs, \ba) \right] \nonumber
+ \lambda (\mathbb{E}_{\bs\sim\mathcal{D}_\text{mix}}\left[ \log \hat{\pi}_{\beta_\text{mix}}(\pi_\theta(\bs)|\bs) \right] - \epsilon_\text{th}) \nonumber 
\end{align} 
Intuitively, this unconstrained objective implies that if the expected log-likelihood is large than a threshold $\epsilon_\text{th}$, $\lambda$ is going to be adjusted to 0, on the contrary, $\lambda$ is likely to take a large value, used to encourage the support constraints, \ie  $\mathbb{E}_{\bs\sim\mathcal{D}_\text{mix}}\left[ \log \hat{\pi}_{\beta_\text{mix}}(\pi_\theta(\bs)|\bs) \right] > \epsilon_\text{th}\label{mixturepolicy}$.

%we learn an ensemble of target transition model $\hat{T}_\text{target}$ to maintain the model's uncertainly and take $1(\cdot > \epsilon)$ over the minimum of ensemble models.

\subsection{Ablation Study}  
To understand how the choice of different components of BOSA affects its performance, we conduct the following ablations: 
\begin{enumerate}
\item (\textit{w/o policy reg.}) We remove the supported policy regularization in Equation~7 (in main paper) to check whether the support constraints (filter operator) in value optimization can accommodate the potential OOD state actions. Specifically, we perform policy optimization with
\begin{align}
\max_{\pi_\theta} \ \mathcal{J}_{\mathcal{D}_\text{mix}}(\pi_\theta) &:= \mathbb{E}_{\bs \sim \mathcal{D}_\text{mix}, \ba\sim\pi_\theta(\ba|\bs)}\left[ Q_\phi(\bs, \ba) \right], \nonumber
\end{align} 
and value optimization with
\begin{align}
\min_{Q_\phi} \ \mathcal{L}_\text{mix}(Q_{{\phi}}):= \mathbb{E}_{^{(\bs, \ba, r, \bs')\sim\mathcal{D}_\text{mix}}_{\ba'\sim\pi_\theta(\ba'|\bs')}} \Big[ \Big. 
\left. \delta(Q_\phi) \cdot \mathbbm{1}( \hat{T}_\text{target}(\bs'|\bs,\ba) > \epsilon_\text{th}')\right] + \mathbb{E}_{(\bs,\ba) \sim \mathcal{D}_\text{source}}\left[ Q_\phi(\bs,\ba) \right]\nonumber
\end{align}
\item (\textit{w/o filter}) We remove the filter operator in value optimization (Equation~8 in main paper) to investigate the cross-domain performance without explicit transition adaptation. 
Specifically, we perform policy optimization with
\begin{align}
\max_{\pi_\theta} \ \mathcal{J}_{\mathcal{D}_\text{mix}}(\pi_\theta) &:= \mathbb{E}_{\bs \sim \mathcal{D}_\text{mix}, \ba\sim\pi_\theta(\ba|\bs)}\left[ Q_\phi(\bs, \ba) \right], \nonumber 
\text{s.t.} \ \mathbb{E}_{\bs\sim\mathcal{D}_\text{mix}}\left[ \log \hat{\pi}_{\beta_\text{mix}}(\pi_\theta(\bs)|\bs) \right] > \epsilon_\text{th}, \nonumber
\end{align} 
and value optimization with
\begin{align}
\min_{Q_\phi} \ \mathcal{L}_\text{mix}(Q_{{\phi}}):= &\mathbb{E}_{(\bs, \ba, r, \bs')\sim\mathcal{D}_\text{mix},\ba'\sim\pi_\theta(\ba'|\bs')} \left[\delta(Q_\phi) \right] \nonumber
+ \mathbb{E}_{(\bs,\ba) \sim \mathcal{D}_\text{source}}\left[ Q_\phi(\bs,\ba) \right]. \nonumber
\end{align}
\item (\textit{w/o conservation}) We remove the conservation regularization (Equation~9 in main paper) in value optimization to check how the usage of conservation affects the performance. 
Specifically, we perform policy optimization with
\begin{align}
\max_{\pi_\theta} \ \mathcal{J}_{\mathcal{D}_\text{mix}}(\pi_\theta) &:= \mathbb{E}_{\bs \sim \mathcal{D}_\text{mix}, \ba\sim\pi_\theta(\ba|\bs)}\left[ Q_\phi(\bs, \ba) \right], \nonumber 
\text{s.t.} \ \mathbb{E}_{\bs\sim\mathcal{D}_\text{mix}}\left[ \log \hat{\pi}_{\beta_\text{mix}}(\pi_\theta(\bs)|\bs) \right] > \epsilon_\text{th}, \nonumber
\end{align} 
and value optimization with
\begin{align}
\min_{Q_\phi} \ \mathcal{L}_\text{mix}(Q_{{\phi}}):= &\mathbb{E}_{(\bs, \ba, r, \bs')\sim\mathcal{D}_\text{mix},\ba'\sim\pi_\theta(\ba'|\bs')} \Big[ \Big. \nonumber
\left. \delta(Q_\phi) \cdot \mathbbm{1}(\log \hat{T}_\text{target}(\bs'|\bs,\ba) > \epsilon_\text{th}')\right]. \nonumber
\end{align}
\item (\textit{w/ target data}) To eliminate the potential OOD transition dynamics, one can directly perform value optimization over the target-domain data. Thus, we replace the expectation of Equation~8 (in main paper) with target-domain data $\mathcal{D}_\text{target}$ and remove the corresponding filter operator. 
Specifically, we perform policy optimization with
\begin{align}
\max_{\pi_\theta} \ \mathcal{J}_{\mathcal{D}_\text{mix}}(\pi_\theta) &:= \mathbb{E}_{\bs \sim \mathcal{D}_\text{mix}, \ba\sim\pi_\theta(\ba|\bs)}\left[ Q_\phi(\bs, \ba) \right], \nonumber
\text{s.t.} \ \mathbb{E}_{\bs\sim\mathcal{D}_\text{mix}}\left[ \log \hat{\pi}_{\beta_\text{mix}}(\pi_\theta(\bs)|\bs) \right] > \epsilon_\text{th}, \nonumber
\end{align} 
and value optimization with
\begin{align}
\min_{Q_\phi} \ \mathcal{L}_\text{mix}(Q_{{\phi}}):= &\mathbb{E}_{(\bs, \ba, r, \bs')\sim\mathcal{D}_\text{target},\ba'\sim\pi_\theta(\ba'|\bs')} \left[ \delta(Q_\phi) \right] \nonumber
+ \mathbb{E}_{(\bs,\ba) \sim \mathcal{D}_\text{source}}\left[ Q_\phi(\bs,\ba) \right]. \nonumber
\end{align}
\end{enumerate}

\subsection{Dataset and Hyper-parameters}
\begin{algorithm}[t]
\caption{Model-based RL data augmentation}
\label{app:alg:dataaug-model}
\textbf{Require:} Target-domain data $\mathcal{D}_\text{target}$. \\
\textbf{Return:} Generated source-domain data $\mathcal{D}_\text{source}$. 

\begin{algorithmic}[1]
    \STATE Initialize source-domain dataset $\mathcal{D}_\text{source} = \varnothing$. 
    \STATE Initialize a pseudo-transition model $\tilde{T}(\bs'|\bs,\ba)$. 
    \FOR{$k=1,\cdots, K$} 
    \STATE  Sample a batch of  data $\mathcal{D}_\text{batch}:=\{ (\bs,\ba,r,\bs') \}_1^n$ from the target-domain dataset {$\mathcal{D}_\text{target}$}. 
    \STATE Optimize $\tilde{T}(\bs'|\bs,\ba)$ by maximizing the log-likelihood $\max_{\tilde{T}} \ \mathbb{E}_{\mathcal{D}_\text{batch}}\left[ \log \tilde{T}(\bs'|\bs,\ba)  \right]$.
    \STATE  Sample a batch of  data $\mathcal{{D'}}_\text{batch}:=\{ (\bs,\ba,r,\bs') \}_1^n$ from the target-domain dataset {$\mathcal{D}_\text{target}$}. 
    \STATE  Generate the next state $\tilde{\bs}'$ by querying the pseudo-model $\tilde{T}(\bs'|\bs,\ba)$, \ie $\tilde{\bs}' \sim \tilde{T}({\bs}'|\bs,\ba)$. 
    \STATE Store the generated transition $(\bs,\ba,r,\tilde{\bs}')$ into $\mathcal{D}_\text{source}$, \\ \ie   $\mathcal{D}_\text{source} \leftarrow \mathcal{D}_\text{source} \cup \{ (\bs,\ba,r,\tilde{\bs}') \}$.
    \ENDFOR
\end{algorithmic}
\end{algorithm}

\begin{algorithm}[t]
\caption{Nosing data augmentation}
\label{app:alg:dataaug-noising}
\textbf{Require:} Target-domain data $\mathcal{D}_\text{target}$ and noise amplitude $s$. \\
\textbf{Return:} Generated source-domain data $\mathcal{D}_\text{source}$. 

\begin{algorithmic}[1]
    \STATE Initialize source-domain dataset $\mathcal{D}_\text{source} = \varnothing$. 
    \FOR{$k=1,\cdots, K$} 
    \STATE  Sample a batch of  data $\mathcal{D}_\text{batch}:=\{ (\bs,\ba,r,\bs') \}_1^n$ from the target-domain dataset {$\mathcal{D}_\text{target}$}. 
    \STATE Sample a batch of noise $\mathbf{n}$, \\ \ie $\mathbf{n} = 2 * (\text{np.random.rand(batch\_size)} - 0.5) * s $. 
    \STATE  Generate next state $\tilde{\bs}'$: $\tilde{\bs}' = \tilde{\bs}' + \mathbf{n}$.
    \STATE Store the generated transition $(\bs,\ba,r,\tilde{\bs}')$ into $\mathcal{D}_\text{source}$, \\ \ie   $\mathcal{D}_\text{source} \leftarrow \mathcal{D}_\text{source} \cup \{ (\bs,\ba,r,\tilde{\bs}') \}$.
    \ENDFOR
\end{algorithmic}

\end{algorithm}

\begin{table}[t]
	\centering
	\caption{Statistics for each task in our cross-domain offline RL tasks. }
 \vspace{2pt}
\label{appendix:environment-statistics-samples}% \textcolor{red}{$\uparrow$}  \textcolor{blue}{$\downarrow$} 
	\begin{adjustbox}{max width=0.95\textwidth}
  \begin{small}
		\begin{tabular}{lllll}
			\toprule 
			\textbf{Environment} & \multicolumn{1}{l}{\textbf{Domain shift}} & \textbf{Task name} & \textbf{Target} (10\%D4RL) & \textbf{Source} \\
			\midrule
			\multicolumn{1}{l}{\multirow{8}[4]{*}{Hopper}} & \multirow{4}[2]{*}{Body mass shfit} & Random & $10^5$ (D4RL) & $10^6$ \\
			&       & Medium & $10^5$ (D4RL) & $10^6$ \\
			&       & Medium-replay & $20092$ (D4RL) & $10^6$ \\
			&       & Medium-expert & $2*10^5$ (D4RL) & $2*10^6$ \\
			\cmidrule{2-5}          & \multirow{4}[2]{*}{Joint noise shift} & Random & $10^5$ (D4RL) & $10^6$ \\
			&       & Medium & $10^5$ (D4RL) & $10^6$ \\
			&       & Medium-replay & $20092$ (D4RL) & $10^6$ \\
			&       & Medium-expert & $2*10^5$ (D4RL) & $2*10^6$ \\
			\midrule
			\multicolumn{1}{l}{\multirow{8}[4]{*}{Walker2d}} & \multirow{4}[2]{*}{Body mass shfit} & Random & $10^5$ (D4RL) & $10^6$ \\
			&       & Medium & $10^5$ (D4RL) & $10^6$ \\
			&       & Medium-replay & $10093$ (D4RL) & $10^6$ \\
			&       & Medium-expert & $2*10^5$ (D4RL) & $2*10^6$ \\
			\cmidrule{2-5}          & \multirow{4}[2]{*}{Joint noise shift} & Random & $10^5$ (D4RL) & $10^6$ \\
			&       & Medium & $10^5$ (D4RL) & $10^6$ \\
			&       & Medium-replay & $10093$ (D4RL) & $10^6$ \\
			&       & Medium-expert & $2*10^5$ (D4RL) & $2*10^6$ \\
			\midrule
			\multicolumn{1}{l}{\multirow{8}[4]{*}{HalfCheetah}} & \multirow{4}[2]{*}{Body mass shfit} & Random & $10^5$ (D4RL) & $10^6$ \\
			&       & Medium & $10^5$ (D4RL) & $10^6$ \\
			&       & Medium-replay & $10100$ (D4RL) & $10^6$ \\
			&       & Medium-expert & $2*10^5$ (D4RL) & $2*10^6$ \\
			\cmidrule{2-5}          & \multirow{4}[2]{*}{Joint noise shift} & Random & $10^5$ (D4RL) & $10^6$ \\
			&       & Medium & $10^5$ (D4RL) & $10^6$ \\
			&       & Medium-replay & $10100$ (D4RL) & $10^6$ \\
			&       & Medium-expert & $2*10^5$ (D4RL) & $2*10^6$ \\
			\midrule
			\multirow{1}[1]{*}{Quadruped robot} & \multirow{1}[1]{*}{Joint noise shift} & Medium-expert & $3*10^4$ (target) & $2*10^6$ (source) \\
			% &       & Medium-Replay & $3*10^4$ (real-world) & $10^6$ (simulator) \\
			% &       & Medium-Expert & $3*10^4$ (real-world) & $2*10^6$ (simulator) \\
			% &       & Medium-Replay-Expert  & $3*10^4$ (real-world) & $2*10^6$ (simulator) \\
			\bottomrule 
		\end{tabular}%
  \end{small}
	\end{adjustbox}
\end{table}

\begin{table}[h]
	\centering
	\caption{Dynamics shift for Hopper, Walker2d, HalfCheetah, and quadruped robot tasks. For the body mass shift, we change the mass of the body in the source MDP. For the joint noise shift, we add a noise (randomly sampling in $[-0.05, +0.05]$) to the actions when we collect the source offline data.}
  \vspace{2pt}
\label{appendix:tab:dynamics-shift-mass-joint}
	\begin{adjustbox}{max width=1.0\textwidth}
 \begin{small}
	\begin{tabular}{lcccc}
		\toprule
		& \multicolumn{2}{c}{Hopper} & \multicolumn{2}{c}{Walker2d} \\
		\cmidrule(r){2-3}  \cmidrule(r){4-5} 
		& \multicolumn{1}{l}{Body mass shfit} & \multicolumn{1}{l}{Joint noise shift} & \multicolumn{1}{l}{Body mass shfit} & \multicolumn{1}{l}{Joint noise shift}  \\
%		\midrule
		Source &  mass[-1]=2.5     &  action[-1]+noise     &  mass[-1]=1.47     &   action[-1]+noise \\
		Target &  mass[-1]=5.0     &  action[-1]+0    &  mass[-1]=2.94     &   action[-1]+0 \\
        \midrule
        & \multicolumn{2}{c}{HalfCheetah} & \multicolumn{2}{c}{Quadruped robot}\\ 
		\cmidrule(r){2-3}  \cmidrule(r){4-5} 
		& \multicolumn{1}{l}{Body mass shfit} & \multicolumn{1}{l}{Joint noise shift} &  \multicolumn{1}{l}{Joint noise shift} \\
%		\midrule
		Source &   mass[4]=0.5    & action[-1]+noise & action+noise\\
		Target &   mass[4]=1.0    & action[-1]+0 & action+0\\
		\bottomrule
	\end{tabular}
  \end{small}
	\end{adjustbox}
\end{table}

\subsubsection{Cross-Domain Offline RL Dataset}
\textbf{Cross-domain tasks.}
For cross-domain offline transfer experiments, we use the D4RL~\cite{d4rl2020Fu} offline data as the target domain and use the collected data from DARA~\cite{liu2022dara} as the source domain (see statistics for each task in Table~\ref{appendix:environment-statistics-samples}). Specifically, the source-domain data are collected by modifying the body {mass} or adding noise to {joints} of the agent (see Table~\ref{appendix:tab:dynamics-shift-mass-joint}) and then following the same data-collection procedure as in D4RL. 

\textbf{Data augmentation.} 
If we can not access additional source-domain data, we can learn a (sub-optimal) pseudo-transition model and use the learned model to generate new transitions (\textit{model-based RL data augmentation, Algorithm~\ref{app:alg:dataaug-model}}). Alternatively, we can also employ (noising) data augmentation to generate new transitions. Then, we can directly treat the generated transitions as the source data (\textit{nosing data augmentation, Algorithm~\ref{app:alg:dataaug-noising}}). 

\subsubsection{Hyper-parameters.}

In Tables~\ref{app:tab:vae-behavior} and~\ref{app:tab:vae-transition}, we provide the hyper-parameters of CVAE implementations of $\hat{\pi}_{\beta_\text{mix}}$ and $\hat{T}_\text{target}$. In Table~\ref{app:tab:hyper-bosa}, we provide the hyper-parameters of our BOSA. 

\begin{table}[htbp]
\caption{Hyper-parameters of CVAE implementation of the cross-domain behavior policy $\hat{\pi}_{\beta_\text{mix}}$.}
\label{app:tab:vae-behavior}
\begin{center}
    \begin{small}
        \begin{tabular}{lll}
            \toprule
            & Hyperparameter       & \multicolumn{1}{l}{Value}               \\ \midrule
            \multirow{7}{*}{CVAE training}     & Optimizer            & \multicolumn{1}{l}{Adam}                \\
            & Learning rate        & $1\times 10^{-3}$                                   \\
            & Batch size           & 256                                     \\
            & Number of iterations & $10^5$                                  \\
            & KL term weight       & 0.5                                     \\
            & Normalized states    & \textit{True} 
            \\
            \midrule
            \multirow{5}{*}{CVAE architecture} & Encoder hidden dim   & 750                                     \\
            & Encoder layers       & 3                                       \\
            & Latent dim           & \multicolumn{1}{l}{2 $\times$ action dim}        \\
            & Decoder hidden dim   & 750                                     \\
            & Decoder layers       & 3                                       \\ \bottomrule
        \end{tabular}
    \end{small}
\end{center}
\end{table}

\begin{table}[htbp]
\caption{Hyper-parameters of CVAE implementation of the target-domain transition dynamics $\hat{T}_\text{target}$.}
\label{app:tab:vae-transition}
\begin{center}
    \begin{small}
        \begin{tabular}{lll}
            \toprule
            & Hyperparameter       & \multicolumn{1}{l}{Value}               \\ \midrule
            \multirow{7}{*}{CVAE training}     & Optimizer            & \multicolumn{1}{l}{Adam}                \\
            & Learning rate        & $1\times 10^{-3}$                                   \\
            & Batch size           & 256                                     \\
            & Number of iterations & $10^5$                                  \\
            & KL term weight       & 0.5                                     \\
            & Normalized states    & \textit{True} 
            \\ 
            \midrule
            \multirow{5}{*}{CVAE architecture} & Encoder hidden dim   & 750                                     \\
            & Encoder layers       & 3                                       \\
            & Latent dim           & \multicolumn{1}{l}{2 $\times$ state dim}        \\
            & Decoder hidden dim   & 750                                     \\
            & Decoder layers       & 3                                       \\ \bottomrule
        \end{tabular}
    \end{small}
\end{center}
\end{table}

%\begin{align}
%   \
%\end{align}

\begin{table*}[htbp]
\caption{Hyper-parameters of our BOSA training. We use TD3 as our base actor-critic RL implementation. }
\label{app:tab:hyper-bosa}
\begin{center}
    \begin{small}
        \begin{tabular}{lll}
            \toprule
            & Hyperparameter          & \multicolumn{1}{l}{Value}           \\ \midrule
            \multirow{11}{*}{TD3}         & Optimizer               & \multicolumn{1}{l}{Adam}            \\
            & Critic learning rate    & \multicolumn{1}{l}{$3\times 10^{-4}$}            \\
            & Actor learning rate     & \multicolumn{1}{l}{$3\times 10^{-4}$} \\
            & Batch size              & 256                                 \\
            & Discount factor                & 0.99                                \\
            & Number of iterations    & $10^6$                             \\
            & Target update rate $\tau$      & 0.005                               \\
            & Policy noise            & 0.2                                 \\
            & Policy noise clipping   & 0.5                                 \\
            & Policy update frequency & 2                                   \\ \midrule
            \multirow{6}{*}{Architecture} & Actor hidden dim        & 256                                 \\
            & Actor layers            & 3                                   \\
            & Actor dropout     & \multicolumn{1}{l}{0.1} \\
            & Critic hidden dim       & 256                                 \\
            & Critic layers           & 3                                   \\ \midrule
            \multirow{5}{*}{BOSA}         
            
            & \multirow{1}{*}{$\lambda_\text{policy}$}                              
            & \multicolumn{1}{l}{$\{0.1, 0.01, 0.05, 0.5, 0.12, 0.15, 0.2\}$ }     \\
            
            & \multirow{1}{*}{$\lambda_\text{transition}$ }                              
            & \multicolumn{1}{l}{$\{0.1, 0.01, 0.05, 0.5, 0.12, 0.15, 0.2\}$ }     \\
            
            & \multirow{1}{*}{ $ \exp{(\epsilon_\text{th})}$ }                              
            & \multicolumn{1}{l}{$\{0.01, 0.1, 0.08, 0.2\}$ }     \\
            & \multirow{1}{*}{$ \exp{(\epsilon_\text{th}')}$ }                              
            & \multicolumn{1}{l}{$\{0.01, 0.1, 0.08, 0.2\}$ }     \\
            
            & \multirow{1}{*}{Amount of CVAEs (ensemble size)}                              
            & \multicolumn{1}{l}{$\{5,1\}$ }     \\
            
            % & \multirow{1}{*}{Filter weight $w_{filter}$ }                              
            % & \multicolumn{1}{l}{$\{ 0.1 \}$ }     \\
            
            & \multirow{1}{*}{Value conservation weight $w_\text{conservation}$ }                              
            & \multicolumn{1}{l}{$\{ 0, 0.1, 0.01 \}$ }     \\
            \bottomrule
        \end{tabular}
    \end{small}
\end{center}
\end{table*}
\newpage

% \subsection{Compute details.} All of our experiments are run on NIVIDA RTX2080Ti GPUs. In particular, because our framework doesn't involve parallel computing or other multi-GPU acceleration methodologies, so increase the amount of GPUs only contributes to run more experiments at the same time rather training speed of one specific experiment.

\subsection{Codebase} 
Our implementation is based on SPOT~\cite{wu2022supported}: https://github.com/thuml/SPOT. We provide our source code in the supplementary material.

\section{More Results}
\subsection{Sensitivity to Target Data Amount and Hyper-parameters}
\begin{figure*}[t]
% \vspace{-2pt}
\begin{center}
\includegraphics[scale=0.25]{figures/amount_hopper_m_joint.pdf}	
\includegraphics[scale=0.25]{figures/amount_HalfCheetah_m_joint.pdf}
\includegraphics[scale=0.25]{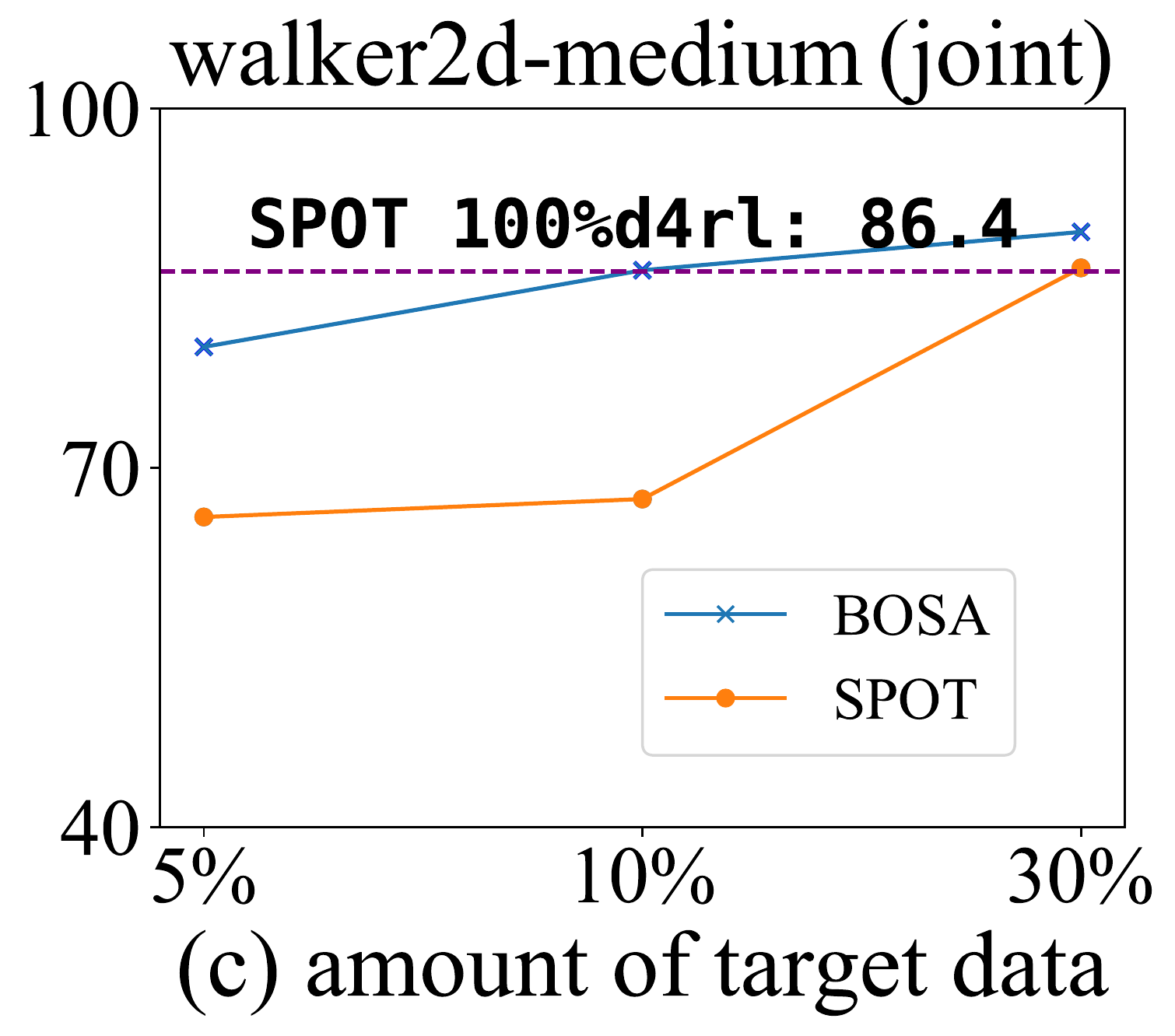} 
\includegraphics[scale=0.25]{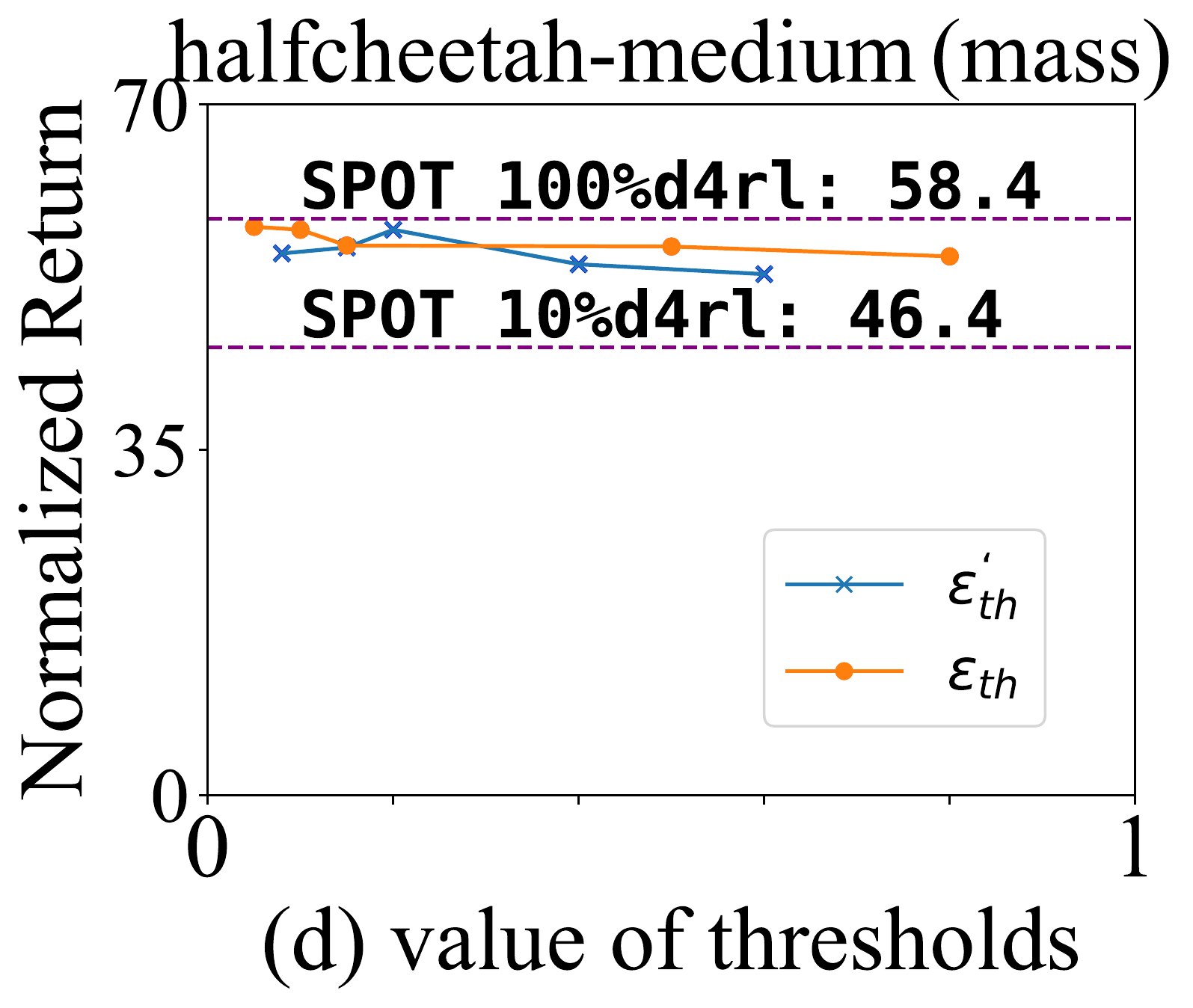}
\includegraphics[scale=0.25]{figures/walker-m-mass-noise.pdf} \ \ \ 
\includegraphics[scale=0.25]{figures/cheetah-m-joint-noise.pdf}
\end{center}
\vspace{-5pt}
\caption{Sensitivity on the amount of target-domain data (a, b, c) and the thresholds $\epsilon_{\text{th}}$ and $\epsilon_{\text{th}}'$ (d, e, f). Across a range of amount of target data and thresholds, BOSA consistently achieves better results compared to SPOT.}
\label{fig:ablation-num-thresholds2}
\vspace{-12pt}
\end{figure*}

In Figure~\ref{fig:ablation-num-thresholds2}, we provide additional experimental results (a, b, c) to investigate the sensitivity of BOSA on the amount of target-domain data and (d, e, f) to study the hyper-parameter sensitivity on the thresholds $\epsilon_\text{th}$ and $\epsilon'_\text{th}$ in the supported policy and value optimization respectively. Consistently, BOSA achieves better performance than SPOT across a wide range of target data and thresholds, showing that BOSA can contribute to the target-domain sample efficiency by using additional cross-domain data. 
% We also provide the concrete numerical results in Tabel~\ref{tab:exp-robustness-test}.

\subsection{BOSA on Cross-domain D4RL.}
\begin{figure*}[h]
%	\vspace{-2pt}
\begin{center}
    \includegraphics[scale=0.25]{figures/figures/introvs-ylabel.pdf}
    \includegraphics[scale=0.25]{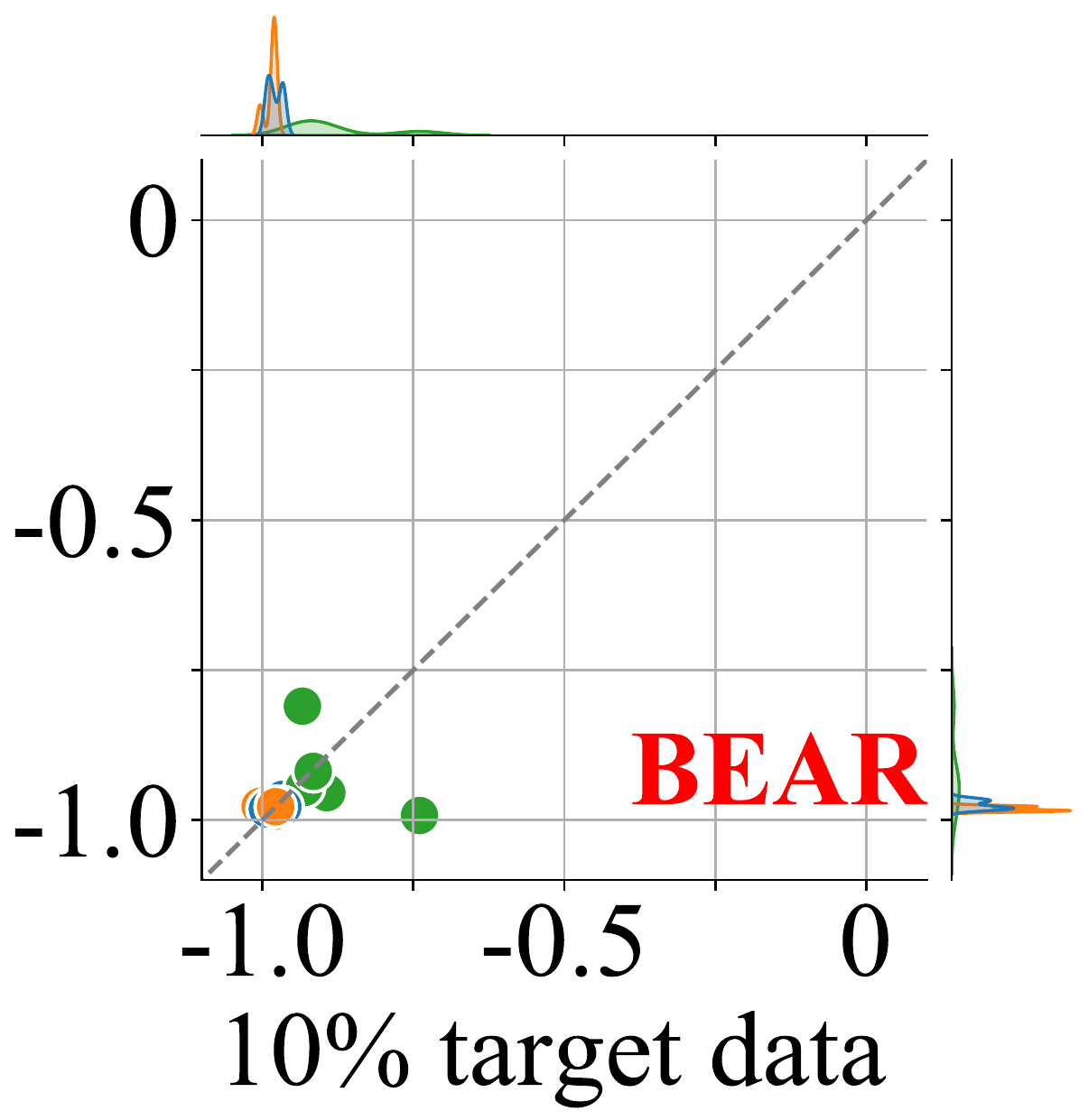}
    \includegraphics[scale=0.25]{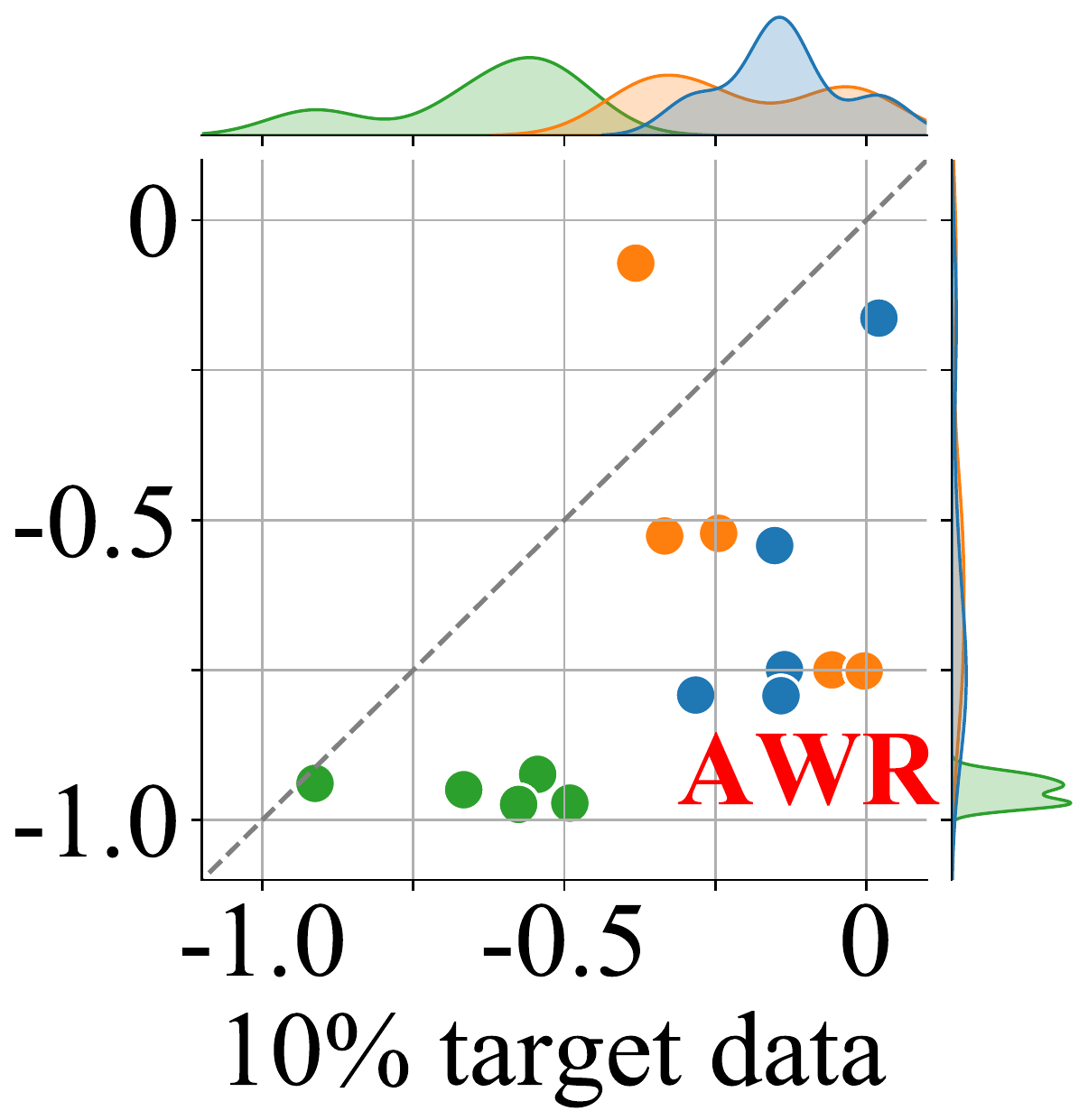}
    \includegraphics[scale=0.25]{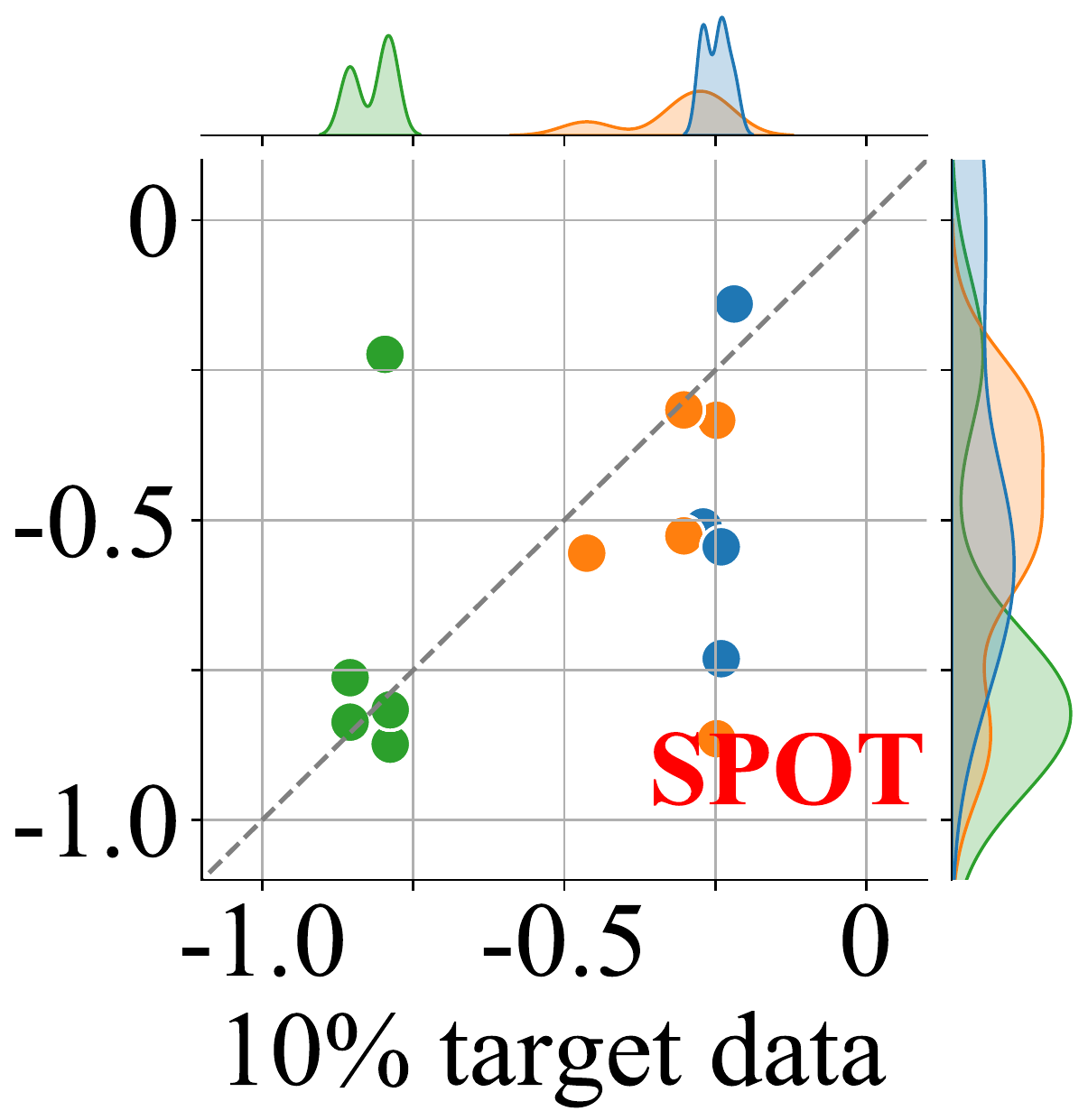}
    \includegraphics[scale=0.25]{figures/figures/introvs-bosa.pdf}
    \includegraphics[scale=0.23]{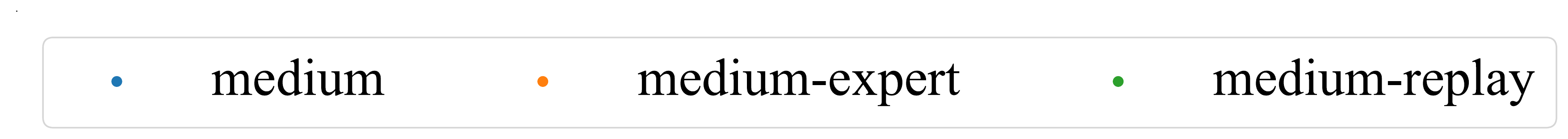}
\end{center}
\vspace{-5pt}
\caption{Performance difference between single-domain and cross-domain offline RL settings, where different colors represent different data qualities (blue: *-medium, green: *-medium-replay, red: *-medium-expert) and multiple dots with the same color represent scores on multiple cross-domain tasks. 
    We take D4RL~\cite{d4rl2020Fu} as the target domain and take a modified D4RL (with transition dynamics shift) as the source domain. The x-axis represents the normalized performance improvement when only 10\% of D4RL data (target) is available, and the y-axis represents the performance difference between learning with cross-domain offline data (10\%target + 100\%source) and learning with abundant target-domain offline data (100\%target). Formally, $\text{x} = \frac{\text{Score}(\text{10\%target})-\text{BestScore}(\text{100\%target})}{\text{BestScore}(\text{100\%target})}$, and  $\text{y} = \frac{\text{Score}(\text{10\%target + 100\%source})-\text{BestScore}(\text{100\%target})}{\text{BestScore}(\text{100\%target})}$. We can observe that when we reduce the offline training data size, most offline RL methods suffer a clear drop in performance (\ie values on the x-axis are less than 0). Further, introducing additional source-domain data also does not bring any significant performance benefits (\ie below the dashed line), with the exception of our BOSA. 
    % (m: medium, mr: medium-replay, me: medium-expert)
    } 
\label{app:fig:intro-vs}
%	\vspace{-2pt}
\end{figure*}

In Figure~\ref{app:fig:intro-vs}, we provide additional results w.r.t the performance comparison between single-domain and cross-domain offline RL settings.
% \end{align}

\subsection{BOSA on Quadruped Robot}

\textbf{Datasets.} We test BOSA on two simulated quadruped robot environments (A1 dog from Unitree). To collect the source-domain data, we add noise to {joints} of the quadruped robot (see Table~\ref{appendix:tab:dynamics-shift-mass-joint}) and then follow the same data-collection procedure as in D4RL to collect the medium-expert-domain dataset.

% \textbf{Baselines.} In terms of our baselines, we selected SPOT, BC\footnote{To be distinguished from traditional BC, we briefly introduce our implementation of behavior cloning(BC). We realize BC with density-estimation which learns an imitation by maximizing the probability of action prediction in action support,\ie $\theta\leftarrow\argmax_{\theta} \pi_{\beta}(\pi_{\theta}(s)|s)$). The motivation of density-BC baseline is to study whether the value function is not an indispensable or indispensable part of policy learning}, and DARA as our baselines. 

\textbf{Demonstration of locomotion states.} We recorded the locomotion states of the quadruped robot by capturing video from the environment render and clipping the resulting video into multiple frames at fixed time intervals (stitched together to form a continuous image). 

%\subfloat[][SPOT, failure，\XSolidBrush]{\includegraphics[scale=0.035]{figures/spot_move.jpg}}
%\subfloat[][DARA, failure，\XSolidBrush]{\includegraphics[scale=0.035]{figures/dara_move.jpg}}
%\subfloat[][BC, failure，\XSolidBrush]{\includegraphics[scale=0.035]{figures/bc_move.jpg}}
\begin{figure*}[ht]
\vspace{-5pt}
\begin{center}
%HalfCheetah-m-Joint-noise.pngspot_label.png
{\includegraphics[scale=0.53]{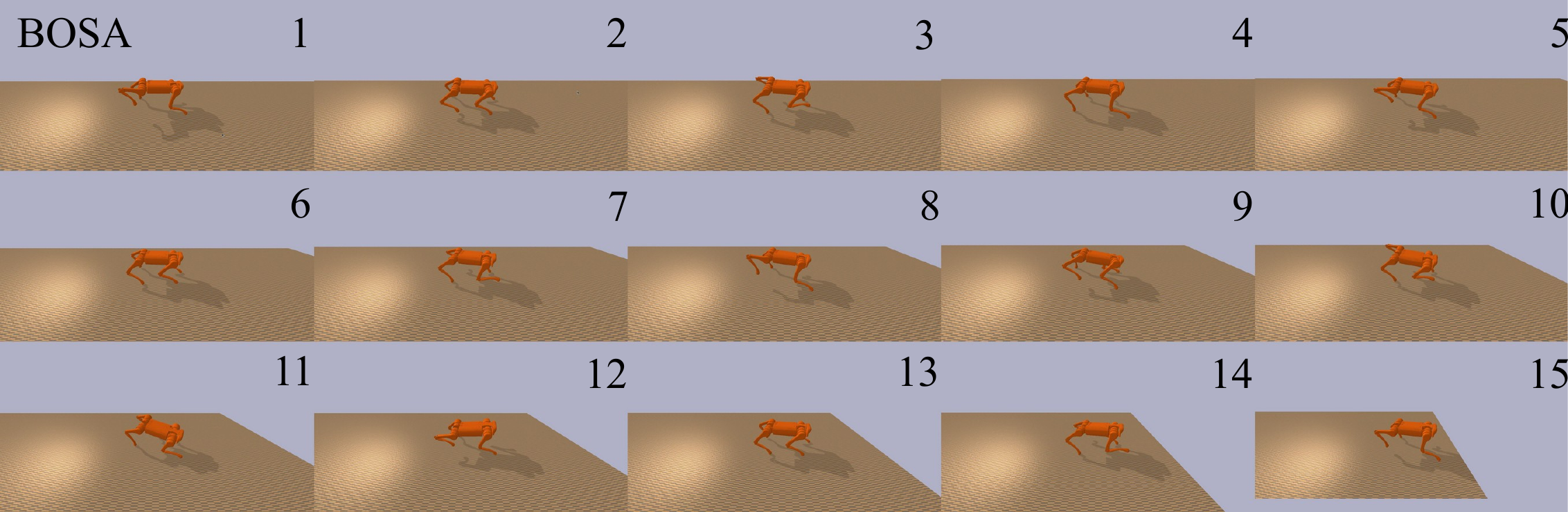}} 
{\includegraphics[scale=0.53]{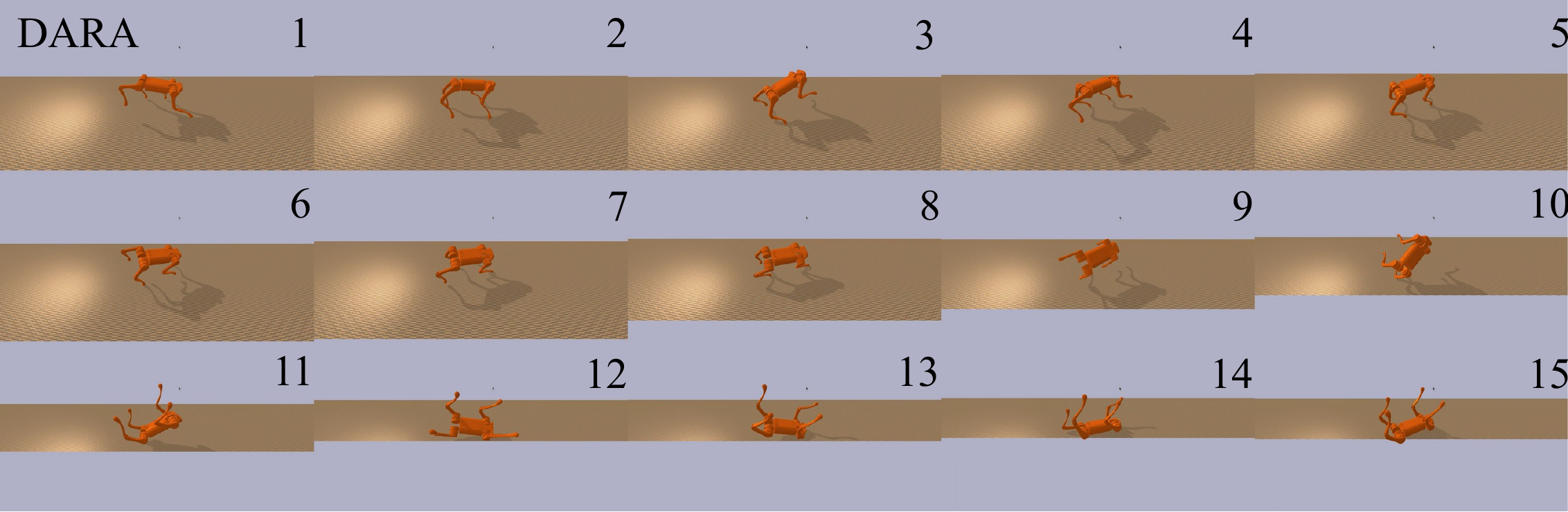}} 
{\includegraphics[scale=0.53]{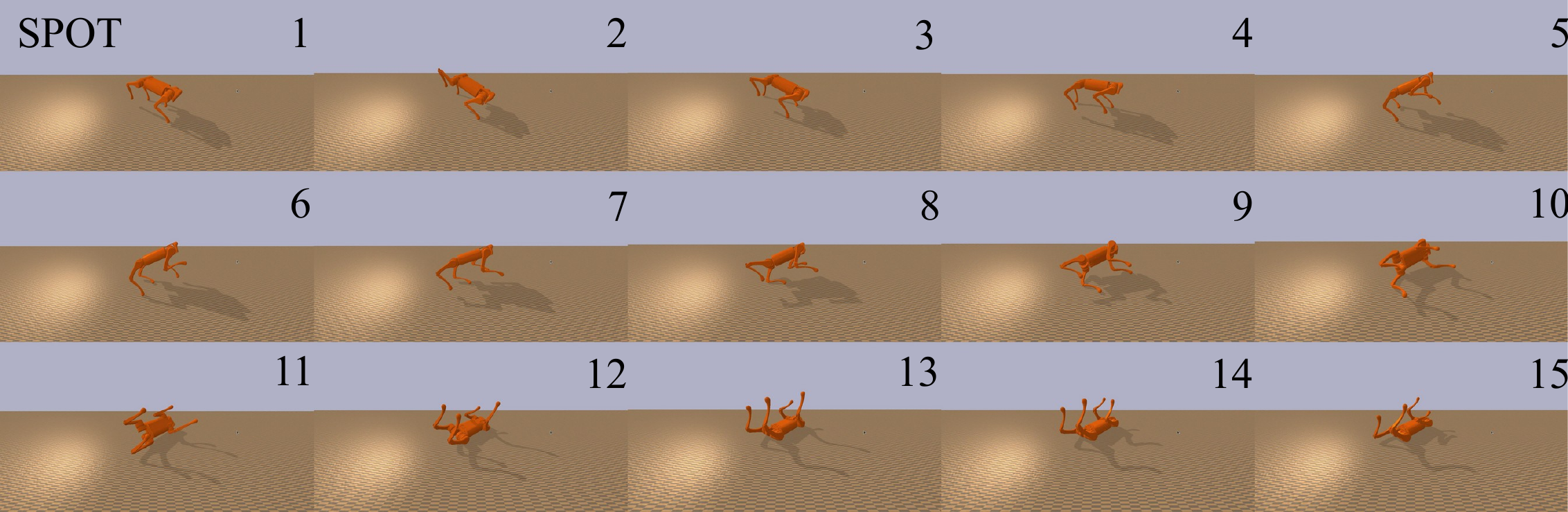}}
{\includegraphics[scale=0.53]{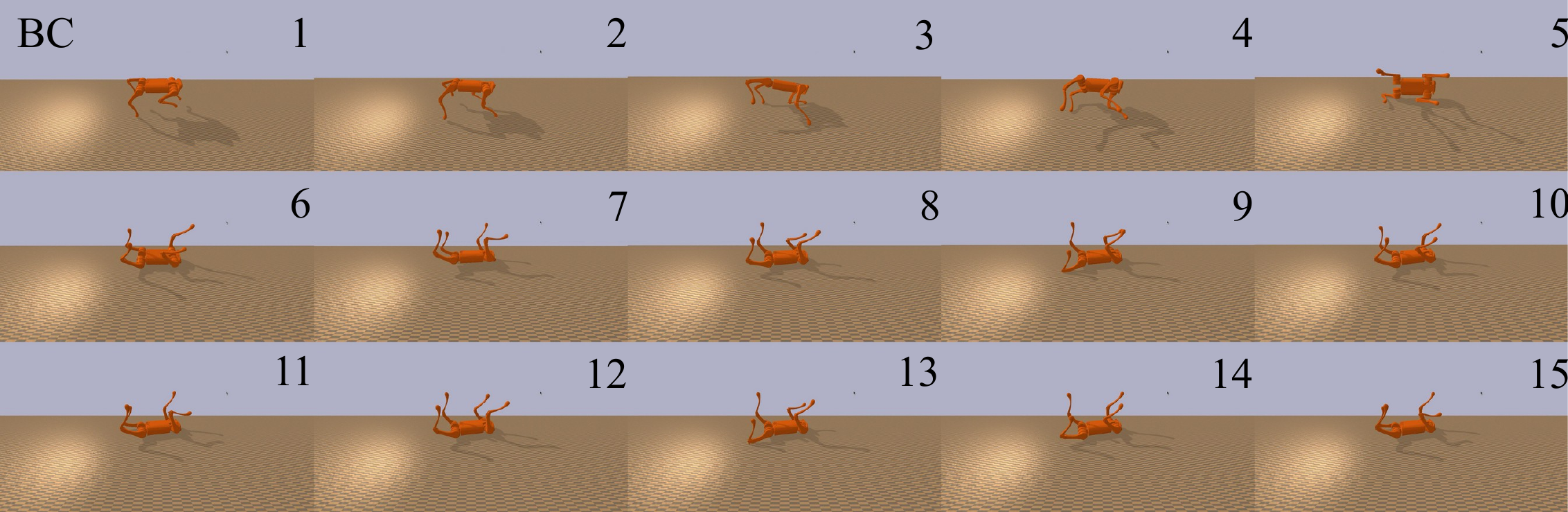}}
\end{center}
\vspace{-4pt}
\caption{Locomotion states of the quadruped robot, where we took 15 images for each baseline. We can find that BOSA achieves significant performance gains in this cross-domain task.}
\label{fig:qutodrap_all_status}
%	\vspace{-2pt}
\end{figure*}

\end{document}